\documentclass{article} 
\usepackage{colm2024_conference}

\usepackage{url}
\usepackage{array}
\usepackage{subfig}
\usepackage{amssymb}
\usepackage{amsmath}
\usepackage{caption}
\usepackage{wrapfig}
\usepackage{booktabs}
\usepackage{graphicx}
\usepackage{hyperref}
\usepackage{multirow}
\usepackage{makecell}
\usepackage{textcomp}
\usepackage{adjustbox}
\usepackage{microtype}
\definecolor{darkblue}{rgb}{0, 0, 0.5}
\hypersetup{colorlinks=true, citecolor=darkblue, linkcolor=darkblue, urlcolor=darkblue}

\title{DCR: Divide-and-Conquer Reasoning for Multi-choice Question Answering with LLMs}

\author{Zijie Meng \thanks{Indicates equal contribution.}, Zhaopeng Feng \& Zuozhu Liu \thanks{Corresponding author.} \\
Zhejiang University-University of Illinois at Urbana Champaign Institute \\
Zhejiang University \\
Jiaxing, Zhejiang 314400, PRC \\
\texttt{\{zijie.22, zhaopeng.23, zuozhuliu\}@intl.zju.edu.cn} \\
\And
Yan Zhang \footnotemark[1] \\
Department of Electrical and Computer Engineering \\
National University of Singapore \\
4 Engineering Drive 3, Singapore 117583 \\
\texttt{yanzhang.jlu@gmail.com}
}

\colmfinalcopy 
\begin{document}
\maketitle

\begin{abstract}
    Large language models (LLMs) have shown impressive performance in reasoning benchmarks with the emergence of Chain-of-Thought (CoT), particularly in multi-choice question (MCQ). However, current works equally resolve questions regardless of the problem-solving difficulty, leading to an excessive focus on simple items while insufficient attention on intricate ones. To address this challenge, we propose a simple yet effective strategy, \textbf{D}ivide and \textbf{C}onquer \textbf{R}easoning (DCR), to enhance the reasoning capability of LLMs for MCQs, as inspired by human beings using heuristics to first categorize tasks and then handle them separately. In particular, we first categorize questions into two subsets based on confidence score $\mathcal{CS}$, which is estimated by statistical frequency of generated answers. Subsequently, we propose Filter Choices based Reasoning (FCR) to improve model performance on MCQs with low $\mathcal{CS}$. Our experiments demonstrate that the proposed strategy only costs 85\% of SOTA, while still achieves average accuracy improvement of 1.56\% across nine datasets including arithmetic, commonsense, and logic reasoning tasks. The code is at \url{https://github.com/AiMijie/DCR}. 
\end{abstract}

\vspace{-4mm}
\section{Introduction}
\vspace{-2mm}
Large language models (LLMs) (e.g., GPT3 \citep{gpt3}, GPT4 \citep{OpenAI2023GPT4TR}, Palm \citep{palm}, Palm2 \citep{palm2}, Lamda \citep{lamda}, Llama \citep{llama}, Llama2 \citep{llama2}) have exhibited outstanding performance on various downstream tasks by generating step by step rationales to obtain final answers without finetuning parameters, as elicited from Chain-of-Thoughts (CoT) \citep{manualCOT}. Multiple-choice question (MCQ) is a format that incorporate a choices list with a question and prompt the model to select the gold answer. Owing to its simple structure, standardized results, and objective assessments, MCQ is not only widely prevalent in the real world but also extensively employed in LLMs' reasoning evaluation \citep{vicuna, mmlu, bigbench, agieval, ceval}. Consequently, the community has witnessed a surge in CoT-based works, which demonstrate outstanding performance on MCQs \citep{self-consistency, activePrompting, zeroShotCOT, PHP, role-play-prompting}. Notably, Zero-Shot-CoT \citep{zeroShotCOT} and Self-Consistency (SC) \citep{self-consistency} have attracted considerable attention due to straightforward implementation and impressive efficacy. Zero-Shot-CoT stimulates the latent zero-shot reasoning abilities of LLMs by adding "Let's think step by step." into prompts, but often underperforms on complex tasks. SC samples different reasoning paths to generate multiple candidates following majority voting to derive final answer, which achieves encouraging results but introduces substantial overhead. To escape this sky-high cost, ESC \citep{ESC} early-stops inference by calculating the entropy of answer distribution in a small sliding window without sacrificing SC's performance, which achieves SOTA currently. However, its performance ceiling is inherently limited by SC, restricting its breakthroughs in accuracy.

Therefore, to optimize the cost and performance, it is imperative to timely halt expensive sampling to reduce expenditure and further employ varied approaches for problems of differing complexity to advance accuracy. In other words, previous methods all process data uniformly regardless of the problem-solving difficulty, which means that simple questions receive unnecessarily complex and costly procedures, whereas intricate ones are not adequately addressed with basic methods. It is also natural that humans utilize heuristic strategies to categorize tasks, and then address each individually, which not only effectively resolves complex issues, but also significantly enhances efficiency \citep{heideman1984gauss, knuth1998sorting}. Consequently, we apply this strategy of data partitioning followed by differential process—Divide and Conquer, which is widely deployed across numerous scenarios \citep{bentley1976divide, bentley1980multidimensional, smith1985design, eisenstein2006divide, mallouk2013divide}—to LLM reasoning. In this context, we need to address two paramount challenges: (1) \textbf{\textit{What criteria should be used to divide the dataset?}} (2) \textbf{\textit{How should the subsets be processed?}}

\begin{figure}[t]
    \centering
    \includegraphics[width=\textwidth]{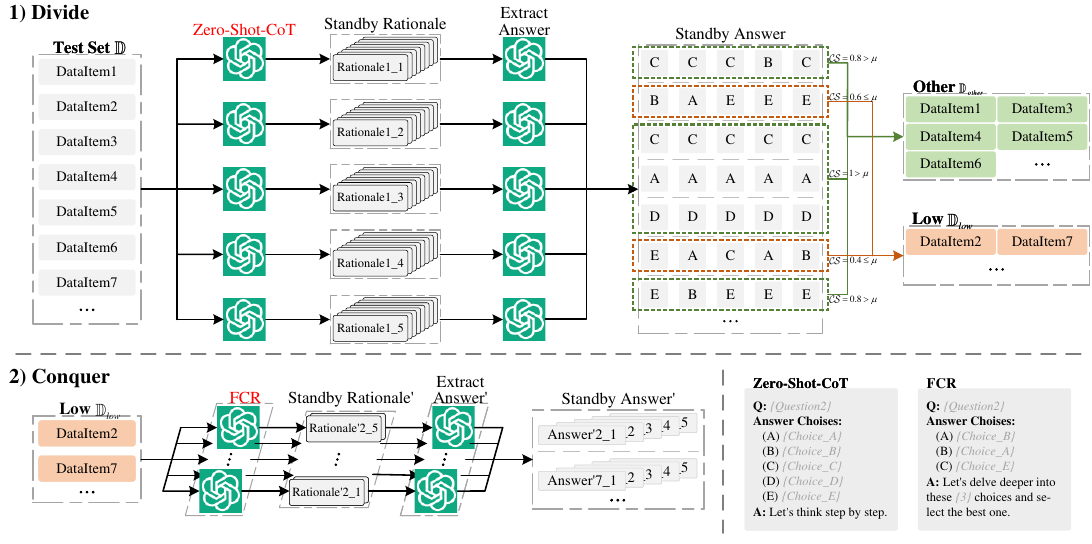}
    \caption{Illustration of DCR. (1) Divide. We first conduct $t$ (e.g. $t$=5) times inference with Zero-Shot-CoT \citep{zeroShotCOT} by ``Let's think step by step.''. Then, the dataset $\mathbb{D}$ is divided based on $\mathcal{CS}$, where DataItems with $\mathcal{CS}$ less than $\mu$ (e.g. $\mu$=0.6) are categorized as $\mathbb{D}_{low}$, and the rest as $\mathbb{D}_{other}$. (2) Conquer. We fix $\mathbb{D}_{other}$ and propose FCR to process $\mathbb{D}_{low}$. ``DataItem'' in Divide area includes question text and full choices list, while involves only filtered choices list in Conquer area. ``Rationale$i$\_$j$'' denotes the rationale generated by $j$-th LLM query for $i$-th DataItem. ``Choice\_$x$'' represents the $x$-th option in original DataItem.}
    \label{fig:pipeline}
    \vspace{-6mm}
\end{figure}

For the first one, we need to explore a method to effectively classify questions based on solving difficulty. In human perception, answers with high uncertainty are often wrong, otherwise tend to be correct \citep{xiong2023llms}. So we tentatively probed SC \citep{self-consistency}, where the statistical distribution of answers generated from various reasoning paths reflects a confidence score $\mathcal{CS}$ for the question. As shown in Figure~\ref{fig:scNumDivide_average_acc}, we divided questions into two subsets based on their $\mathcal{CS}$, where different subsets displays distinct accuracy and the subset with lower $\mathcal{CS}$ demonstrates poorer performance. This suggests that we can employ SC to compute $\mathcal{CS}$ for each problem and divide them.

Move to the second issue, we inspired from the Cannikin Law in management \citep{goldratt2016goal}, explore more elaborately designed methods for the low confidence subsets that offer greater room for optimization, and fix other questions that are sufficiently simple for the model. \citet{irrelevant} investigated the model’s sensitivity to irrelevant information within the questions, but there exists uncertainty regarding irrelevant options in choices list. To delve into this problem, we conducted preliminary studies as shown in Figure~\ref{fig:ac_choiceNum_cmp}, discovering a decrease in problem-solving accuracy as the number of choices increased. Following this, we removed some irrelevant options in hardly solved subsets to re-query the LLM, resulting in a universal improvement of over 20\%, especially achieving staggering 75.52\% on CMSQA \citep{commonsenseqa}, as shown in Table~\ref{tab:choicesNum_effect}. Motivated by these findings, we introduce Filter Choices based Reasoning (FCR), which excludes abundant options by using the answers from the divide stage, to conduct inference in conquer stage.

Concretely, in this paper, we propose a simple yet effective strategy, \textbf{D}ivide and \textbf{C}onquer \textbf{R}easoning (DCR), which first categorizes questions into two subsets based on $\mathcal{CS}$ and subsequently employs FCR to improve model performance on MCQs with low $\mathcal{CS}$, as illustrated in Figure~\ref{fig:pipeline}. Through extensive empirical evaluation across nine datasets including arithmetic, commonsense, and logic tasks, DCR not only consumes on average only 85\% of resources required by ESC, but also improves accuracy by an average of 1.56\% on these datasets. Additionally, we have validated the effectiveness of DCR across various LLMs \citep{team2024gemma, jiang2023mistral, palm2, team2023gemini, OpenAI2023GPT4TR} and the superiority of FCR over other reasoning methods \citep{manualCOT, activePrompting, PHP, zeroShotCOT, role-play-prompting}. We have also successfully adapted DCR to the cloze-style dataset GSM8K \citep{gsm8k} achieving an improved performance over SC with reduced cost. In summary, our work has three major contributions: (1) To the best of our knowledge, we pioneeringly employ the Divide and Conquer at the dataset level for LLM reasoning, providing the community a fresh perspective. (2) By dividing dataset based on $\mathcal{CS}$ and conquering low $\mathcal{CS}$ subset with FCR, we achieve an optimal balance between cost and accuracy. (3) We evaluate this strategy across nine datasets within three distinct reasoning tasks, consistently yielding significant improvements.

\vspace{-2mm}
\section{Methodology}
\vspace{-2mm}

The overall framework of DCR is illustrated in Figure~\ref{fig:pipeline}. Given a test set of length $n$ represented as $\mathbb{D}=\{(Q_1, \mathbf{C}_1), ... , (Q_n, \mathbf{C}_n)\}$, where $Q_i$ denotes the $i$-th question text and $\mathbf{C}_i$ is the corresponding choices list. In addition, we use $\mathbf{R}_i$ and $\mathbf{A}_i$ to denote its rationales and answers generated by LLMs, respectively.

\vspace{-2mm}
\subsection{Divide} \label{sec:method_divide}
\vspace{-2mm}
With each item $(Q_i, \mathbf{C}_i), i\in\{1, ..., n\}$, we query the LLM for $t$ times to obtain rationales $\mathbf{R}_i=\{r_{i, 1}, ..., r_{i,t}\}$ and corresponding standby answers $\mathbf{A}_i=\{a_{i,1}, ..., a_{i,t}\}$ based on Zero-Shot-CoT\footnote{Fow-Shot-CoT (i.e. CoT) \citep{manualCOT} requires substantial human labor to annotate task-specific examplars, and zero-shot gradually approaches or even surpasses few-shot as the scale of model increases \citep{code-prompting, agieval}. See Table~\ref{tab:zt_ft_cmp} in Appendix~\ref{app:zt_ft_cmp} for our verification.} \citep{zeroShotCOT}. In particular, we set $t$ generally equal to the length of choices list $|\mathbf{C}_i|$, considering the worst scenario where all choices could be sampled. And we conduct a more detailed analysis on different values of $t$ in Section~\ref{sec:study_for_divide}. We use $\delta$ to denote LLM and define the confidence score $\mathcal{CS}$ of each item as:
\begin{equation}
    \mathcal{CS}_{(Q_i, \mathbf{C}_i)}=\max_{j\in\{1, ..., t\}} p(a_{i,j}|\delta(Q_i, \mathbf{C}_i)),
\end{equation}
where $p(a_{i,j}|\delta(Q_i, \mathbf{C}_i))$ is the frequency of $a_{i,j}$ in all predicted answers. And we define it as:
\begin{equation}
    p(a_{i,j}|\delta(Q_i, \mathbf{C}_i))=\frac{\sum_{k\in\{1,...,t\}}\mathbf{1}_{a_{i,k}=a_{i,j}}}{t}.
\end{equation}

Intuitively, $\mathcal{CS}$ indicates the proportion of the most frequent answer among all predicted results in $t$ times inferences, which is employed to reflect the problem-solving difficulty. Then, we can divide $\mathbb{D}$ with the following rule:
\begin{equation}
    (Q_i, \mathbf{C}_i) \in 
    \begin{cases}
        \mathbb{D}_{other}, & \text{if } \mathcal{CS}_{(Q_i, \mathbf{C}_i)} > \mu, \\
        \mathbb{D}_{low}, & \text{if } \mathcal{CS}_{(Q_i, \mathbf{C}_i)} \leq \mu,
    \end{cases}
\end{equation}
where $\mathbb{D}_{low}$ represents the low confidence subset containing $(Q_i, \mathbf{C}_i)$ with dispersed distribution of $\mathbf{A}_i$ and $\mathbb{D}_{other}$ includes rest items. $\mu$ is the threshold for dividing, which is specified in Section~\ref{sec:implementation} and discussed in Section~\ref{sec:study_for_divide}. Moreover, different from dividing the questions into two subsets, we explore a more fine-grained division in Section~\ref{sec:study_for_divide} to evaluate our dividing rule. Next, we would fix $\mathbb{D}_{other}$ to conserve resources, while delve deeper into $\mathbb{D}_{low}$ for ongoing performance improvement.

\vspace{-3mm}
\subsection{Conquer} \label{sec:method_conquer}
\vspace{-2mm}
We propose Filter Choices based Reasoning (FCR) to conquer $\mathbb{D}_{low}$ in this stage, as shown in Figure~\ref{fig:pipeline}, where we exploit the results obtained in divide phase as alternative options for subsequent inference. Specifically, we define $\mathbf{C}_i'=uniq(\mathbf{A}_i)$, where the $uniq(\cdot)$ operation signifies deduplication of $\mathbf{A}_i=\{a_{i,1}, ..., a_{i,t}\}$. Then we use $(Q_i, \mathbf{C}_i')$ to construct the new prompt and query the LLM with ``\textcolor{gray}{Let’s delve deeper into these \{\textit{$|\mathbf{C}_i'|$}\} choices and select the best one.}''\footnote{We evaluate the robustness of FCR for different query prompts in Appendix~\ref{app:diffPrompt}.}. Subsequently, through additional inference for $t$ times, we obtain the new standby answers $\mathbf{A}_i'=\{a'_{i,1}, ..., a'_{i,t}\}$ for $\mathbb{D}_{low}$. Notably, our method does not merely delete options, rather it involves a synchronous modification of the option symbols (i.e. `A', `B', `C', etc.) based on the number of remaining choices. Furthermore, in Section~\ref{sec:study_for_conquer}, we evaluate the impact of conquering different subsets, and compare FCR with other reasoning methods, to demonstrate the superiority of only processing $\mathbb{D}_{low}$ with our method.

Ultimately, we align the standby answers $\{\mathbf{A}_i|(Q_i, \mathbf{C}_i)\in\mathbb{D}_{other}\}$ and $\{\mathbf{A}_i'|(Q_i, \mathbf{C}_i)\in\mathbb{D}_{low}\}$ generated in different stage with $\mathbb{D}_{other}$ and $\mathbb{D}_{low}$, respectively, then utilize majority voting \citep{self-consistency} to determine the final answer for each data item. It is evident that our full strategy requires no human intervention or manual labor, and infers $t$ times for each data item in $\mathbb{D}_{other}$ and $2t$ times for $\mathbb{D}_{low}$.

\vspace{-3mm}
\section{Experiments}
\vspace{-2mm}
\subsection{Datasets and evaluation metrics}
\vspace{-2mm}
To evaluate the effectiveness and empirically analyse DCR, we conducted experiments on three tasks: 1) Arithmetic. AQuA (AQ.) \citep{aqua} and Abstract Algebra (Alg.), High School Mathematics (Math.) from the MMLU dataset \citep{mmlu}. 2) Commonsense. CMSQA (CMS.) \citep{commonsenseqa}, OpenBookQA (OB.) \citep{openbookqa} and ARC Challenge (ARC.) \citep{arc}. 3) Logic. RiddleSense (Rid.) \citep{riddlesense}, Logical Deduction (Logi.) from BIG-bench dataset \citep{bigbench} and Reclor (Rec.) \citep{reclor}. The statistical details can be found in Table~\ref{tab:dataset} of appendix. Additionally, we employed exact match (EM) accuracy to evaluate the performance, which is same as previous works \citep{manualCOT, zeroShotCOT}.

\vspace{-3mm}
\subsection{Implementation details} \label{sec:implementation}
\vspace{-3mm}
We primarily employed $\text{GPT-3.5-Turbo-0613}$ from OpenAI API\footnote{\url{https://platform.openai.com}}, and conducted experiments on other opensource and blackbox LLMs in Section~\ref{sec:cmp_llms}. During the divide phase, we set the temperature to 0.7, and set inference times $t$ to 4 or 5 for different datasets, as detailed in Table~\ref{tab:dataset}. We divided each dataset into $\mathbb{D}_{other}$ and $\mathbb{D}_{low}$ with $\mu$ as 0.6. In the conquer stage, the temperature and inference times were consistent with previous phase. Experiments were conducted on the full dataset by default unless in Section~\ref{sec:study_irrchoices} and Appendix~\ref{app:zt_ft_cmp}, where we randomly sampled 500 items for each dataset except 254 for AQuA and 300 for SVAMP. In addition, the final results were all obtained by averaging five random trials. Notably, considering the accuracy for ESC normally equals to or underperforms SC, we mainly compared with SC in Section~\ref{sec:analyse}.

\vspace{-3mm}
\subsection{Main results} \label{sec:main_result}
\vspace{-2mm}
\begin{table}[t]
    \centering
    \caption{Comparison of problem-solving accuracy (\%) among different methods. ``Avg.'' denotes average accuracy across nine datasets. ``\#Call'' refers to the average sample size (\textbf{i.e. inference times}) for each question across nine datasets. SC$^{*}$ and ESC$^{*}$ represent the versions with approximate sample size of DCR.}
    \resizebox{\textwidth}{!}{%
    \label{tab:mainResult}
    \begin{tabular}{c|ccccccccc|cc}
         \toprule
         \multirow{2.5}{*}{Method} & \multicolumn{3}{c}{Arithmetic} & \multicolumn{3}{c}{Commonsense} & \multicolumn{3}{c|}{Logic} & \multirow{2.5}{*}{\makecell{Avg.}} & \multirow{2.5}{*}{\#Call} \\
         \cmidrule(r{1mm}){2-4} \cmidrule(lr{1mm}){5-7} \cmidrule(l{1mm}){8-10}
         {} & {AQ.} & {Alg.} & {Math.} & {CMS.} & {OB.} & {ARC.} & {Rid.} & {Logi.} & {Rec.} & {} & {} \\
         \midrule
         {SC} & {68.98} & {43.20} & {64.00} & {76.12} & \textbf{{87.04}} & {89.68} & {68.72} & {48.07} & {61.84} & {67.52} & {8.94} \\
         {ESC} & {68.98} & {43.20} & {64.00} & {76.12} & \textbf{{87.04}} & {89.68} & {68.72} & {48.07} & {61.84} & {67.52} & {6.79} \\
         {DCR} & {\textbf{71.02}} & {\textbf{48.60}} & {\textbf{66.52}} & {\textbf{77.97}} & {86.80} & {\textbf{89.79}} & {\textbf{68.81}} & {\textbf{50.27}} & {\textbf{61.96}} & {\textbf{69.08}} & {\textbf{5.79}} \\
         \midrule
         {SC$^{*}$} & {66.46} & {43.20} & {62.52} & {75.00} & {85.24} & {88.98} & {68.03} & {48.80} & {61.00} & {66.58} & {6.17} \\
         {ESC$^{*}$} & {68.98} & {42.20} & {64.00} & {76.12} & {84.68} & {88.52} & {68.72} & {48.07} & {60.20} & {66.83} & {6.17} \\
         \bottomrule 
    \end{tabular}%
    }
\end{table}

We took a comparison between SC \citep{self-consistency}, ESC \citep{ESC}, and our method across nine datasets, as shown in Table~\ref{tab:mainResult}. According to Section~\ref{sec:method_conquer}, we set $2t$ as the upperbound of inference times and defined the window size of ESC as $t$. Upon achieving this limitation, the average sample size (i.e. inference times) for each question of original SC is 8.94 with average accuracy as 67.52\%. ESC reduces the sample size to 6.79 and maintains the accuracy of 67.52\%. DCR further reduces the average sample size to 5.79 while achieves the accuracy of 69.08\%, surpassing baselines with 1.56\%, which demonstrates dual improvements in efficiency and performance.

\begin{wrapfigure}{r}{.4\textwidth}
    \vspace{-4mm}
    \centering
    \includegraphics[width=.4\textwidth]{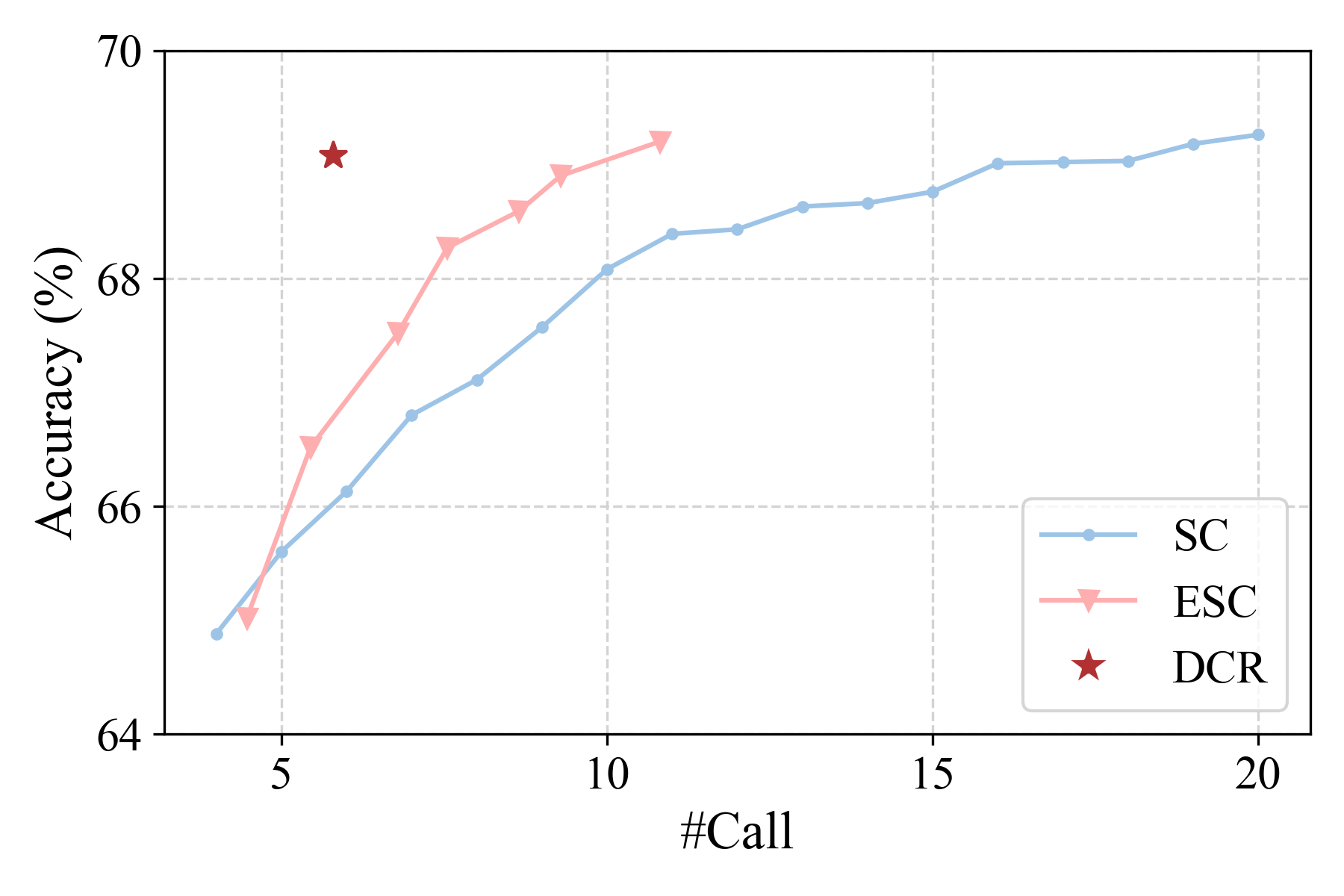}
    \caption{Average accuracy and \#Call across different datasets. See Figure~\ref{fig:trend_ac_cost_detail} for details about each dataset.}
    \label{fig:trend_ac_cost}
\end{wrapfigure}

In Figure~\ref{fig:trend_ac_cost}, we presented the average accuracy of SC and ESC across different datasets for various sample sizes. Notably, DCR achieves similar levels of accuracy at a substantially lower cost compared to the baselines, indicating a significant enhancement in efficiency.  Meanwhile, when costs are comparable, DCR consistently outperforms these two baselines. This superiority is quantitatively reported as SC$^{*}$ and ESC$^{*}$ in Table~\ref{tab:mainResult}, where DCR exhibits an encouraging lead of 2.5\% and 2.25\%, respectively. Furthermore, we observed a diminishing performance improvement of SC and ESC as sample size increases, suggesting an approach towards a bottleneck. However, the integration of FCR for inference on $\mathbb{D}_{low}$ during the conquer stage offers a potential pathway to breakthrough beyond this bottleneck.

\subsection{Analysis} \label{sec:analyse}
\subsubsection{Comparison across different LLMs} \label{sec:cmp_llms}
\begin{wraptable}{r}{.44\textwidth}
    \vspace{-20mm}
    \centering
    \caption{Accuracy (\%) across different LLMs. The number in parenthesis denotes average sample size.}
    \label{tab:diffModel}
    \resizebox{.44\textwidth}{!}{%
    \begin{tabular}{cc|cc}
        \toprule
        \multicolumn{2}{c|}{Setting} & {AQ.} & {CMS.} \\
        \midrule
        \multirow{2}{*}{Gemma} & {SC} & {34.96 (8.00)} & {65.31 (6.00)} \\
        {} & {DCR} & {\textbf{37.24 (7.50)}} & \textbf{{67.81 (5.98)}} \\
        \midrule
        \multirow{2}{*}{Mistral} & {SC} & {39.29 (9.00)} & {71.37 (7.00)} \\
        {} & {DCR} & \textbf{{43.31 (8.97)}} & \textbf{{73.10 (6.51)}} \\
        \midrule
        \multirow{2}{*}{Palm2} & {SC} & {38.50 (6.00)} & {74.15 (6.00)} \\
        {} & {DCR} & \textbf{{39.37 (5.21)}} & \textbf{{75.17 (5.53)}} \\
        \midrule
        \multirow{2}{*}{Gemini} & {SC} & {\textbf{70.39} (8.00)} & {78.41 (6.00)} \\
        {} & {DCR} & {68.74 \textbf{(7.40)}} & \textbf{{78.85 (5.26)}} \\
        \midrule
        \multirow{2}{*}{GPT4} & {SC} & {84.17 (6.00)} & {84.21 (6.00)} \\
        {} & {DCR} & {\textbf{85.43 (5.99)}} & {\textbf{85.19 (5.70)}} \\
        \bottomrule
    \end{tabular}%
    }
\end{wraptable}

In this section, we conducted a comparative analysis between SC \citep{self-consistency} and DCR using various models. Specifically, we employed Gemma (gemma-7b-it) \citep{team2024gemma} and Mistral (Mistral-7B-Instruct-v0.2) \citep{jiang2023mistral} available on the Hugging Face\footnote{\url{https://huggingface.co}}, Palm2 (text-bison-001) \citep{palm2} and Gemini (gemini-pro) \citep{team2023gemini} from Google AI\footnote{\url{https://ai.google.dev}}, as well as GPT4 (gpt-4-1106-preview) \citep{OpenAI2023GPT4TR} from OpenAI API. As shown in Table~\ref{tab:diffModel}, DCR generally achieves higher accuracy with lower costs, except on AQuA using Gemini. Notably, the larger-scale LLMs (e.g. Gemini and GPT4) significantly outperforms other models, particularly on AQuA with improvements exceeding 30\%. However, this also diminishes the relative advantage from DCR, such as the improvements with Mistral are 4.02\% and 1.73\% on two datasets, while only 1.26\% and 0.98\% with GPT4. Therefore, we believe the enhancement of model capabilities resembling the process of making up for weaknesses, which compresses the space for optimization.

\subsubsection{Study for divide stage} \label{sec:study_for_divide}
\begin{figure}[t]
  \centering
  \begin{minipage}[c]{0.49\textwidth}
    \includegraphics[width=\textwidth]{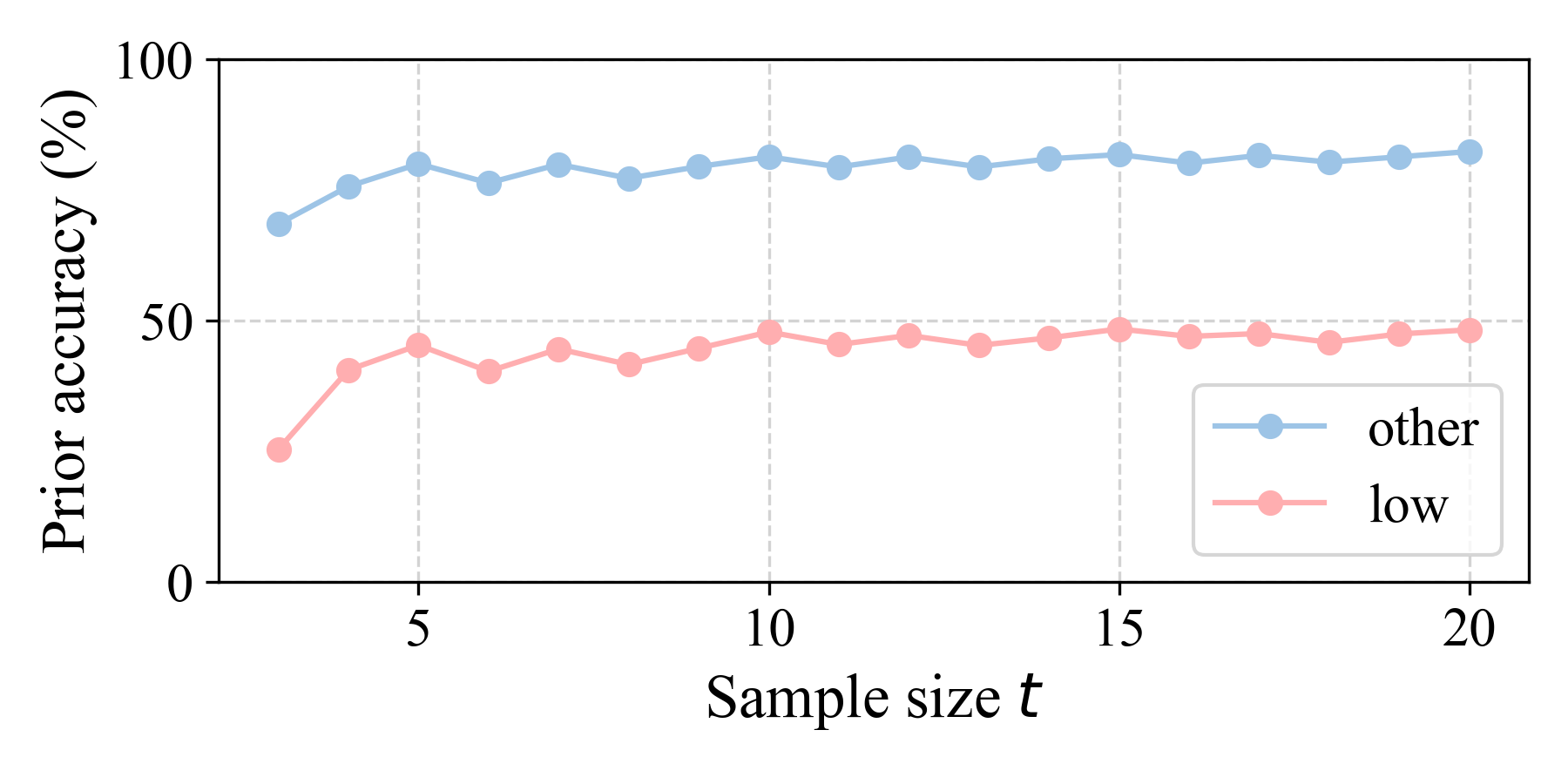}
    \caption{Average Prior accuracy on different subsets for various sample size $t$. See Figure~\ref{fig:scNumDivide_detail_acc} for details about each dataset.}
    \label{fig:scNumDivide_average_acc}
  \end{minipage}
  \hfill
  \begin{minipage}[c]{0.49\textwidth}
    \includegraphics[width=\textwidth]{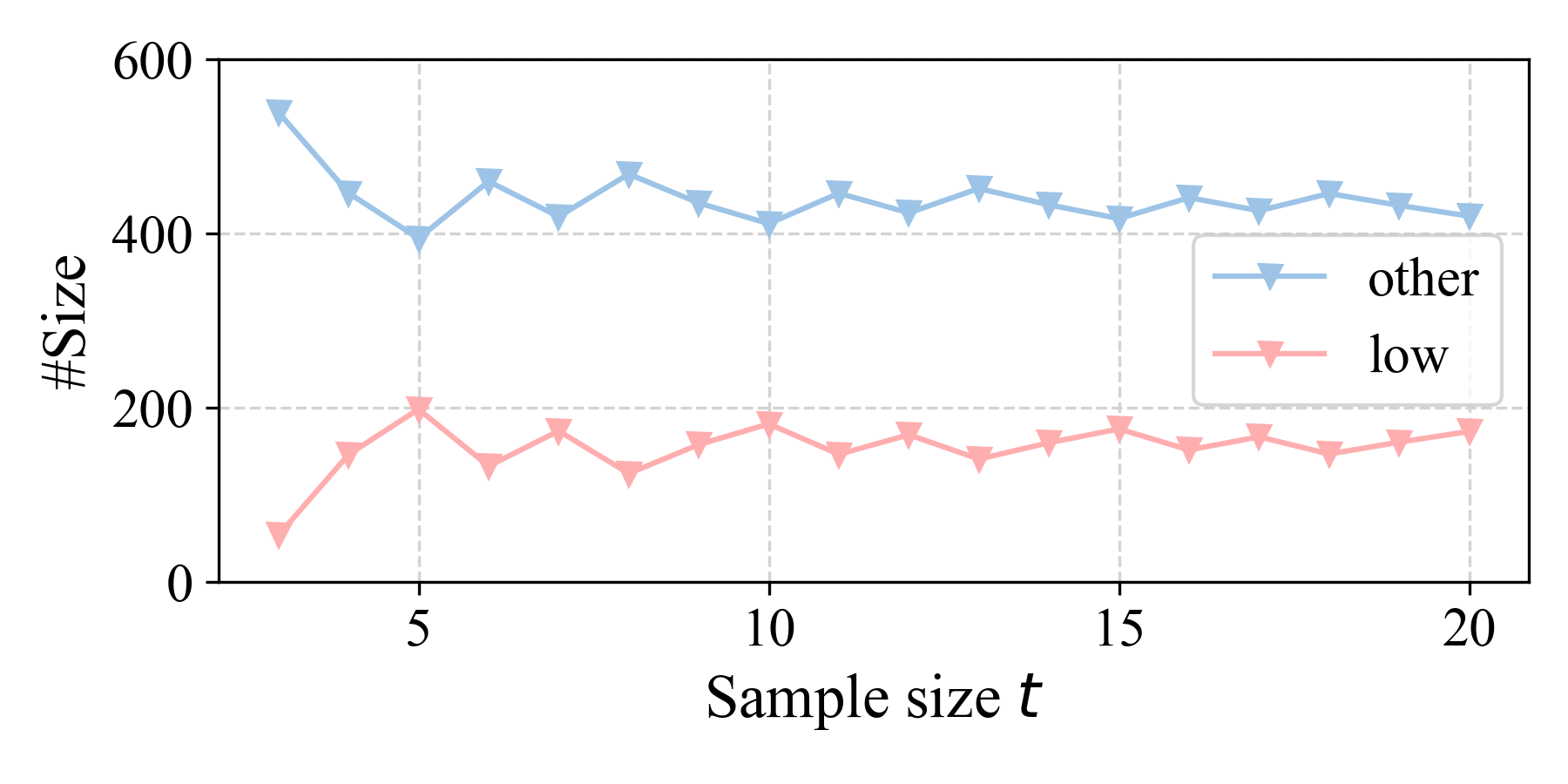}
    \caption{The average number of different subsets size for various sample size $t$. See Figure~\ref{fig:scNumDivide_detail_size} for details about each dataset.}
    \label{fig:scNumDivide_average_size}
  \end{minipage}
\end{figure}

\begin{table*}[t]
    \centering
    \small
    \caption{Comparison of accuracy (\%) on different confidence subsets. ``\textcolor{gray}{\#Size}'' indicates the number of data items in different subsets. ``Prior'' denotes the accuracy of results generated in the divide stage. For $\mathbb{D}_{low_{b}}$, ``-'' refer results lacking reliability because of insufficient data.}
    \label{tab:diffSubset}
    \resizebox{\textwidth}{!}{%
    \begin{tabular}{cc|ccccccccc|c}
         \toprule
         \multirow{2.5}{*}{Subset} & \multirow{2.5}{*}{Setting} & \multicolumn{3}{c}{Arithmetic} & \multicolumn{3}{c}{Commonsense} & \multicolumn{3}{c|}{Logic} & \multirow{2.5}{*}{\makecell[c]{Avg.}} \\
         \cmidrule(r{1mm}){3-5} \cmidrule(lr{1mm}){6-8} \cmidrule(l{1mm}){9-11}
         {} & {} & {AQ.} & {Alg.} & {Math.} & {CMS.} & {OB.} & {ARC.} & {Rid.} & {Logi.} & {Rec.} & {} \\
         \midrule
         \multirow{2.5}{*}{$\mathbb{D}_{high}$} & {\textcolor{gray}{\#Size}} & {\textcolor{gray}{74.60}} & {\textcolor{gray}{30.40}} & {\textcolor{gray}{71.80}} & {\textcolor{gray}{588.80}} & {\textcolor{gray}{325.80}} & {\textcolor{gray}{856.20}} & {\textcolor{gray}{404.40}} & {\textcolor{gray}{27.60}} & {\textcolor{gray}{232.60}} & {\textcolor{gray}{290.24}} \\
         \cmidrule{2-12}
         {} & {Prior} & {91.96} & {53.95} & {90.53} & {92.09} & {96.13} & {96.26} & {89.81} & {86.96} & {75.41} & {85.90} \\
         \midrule
         \multirow{3.5}{*}{$\mathbb{D}_{med}$} & {\textcolor{gray}{\#Size}} & {\textcolor{gray}{51.20}} & {\textcolor{gray}{38.20}} & {\textcolor{gray}{82.40}} & {\textcolor{gray}{265.60}} & {\textcolor{gray}{97.00}} & {\textcolor{gray}{191.20}} & {\textcolor{gray}{218.40}} & {\textcolor{gray}{81.00}} & {\textcolor{gray}{151.40}} & {\textcolor{gray}{130.71}} \\
         \cmidrule{2-12}
         {} & {Prior} & {\textbf{79.69}} & {\textbf{43.46}} & {61.17} & {\textbf{72.74}} & {\textbf{73.61}} & {\textbf{75.42}} & {\textbf{69.60}} & \textbf{{60.49}} & {51.52} & {\textbf{65.30}} \\
         {} & {FCR} & {74.22} & {35.08} & {\textbf{62.86}} & {69.95} & {70.31} & {73.95} & {63.00} & {53.58} & {\textbf{52.84}} & {61.75} \\
         \midrule
         \multirow{3.5}{*}{$\mathbb{D}_{low_{t}}$} & {\textcolor{gray}{\#Size}} & {\textcolor{gray}{70.00}} & {\textcolor{gray}{30.20}} & {\textcolor{gray}{109.60}} & {\textcolor{gray}{265.20}} & {\textcolor{gray}{74.00}} & {\textcolor{gray}{115.40}} & {\textcolor{gray}{251.40}} & {\textcolor{gray}{153.40}} & {\textcolor{gray}{113.20}} & {\textcolor{gray}{131.38}} \\
         \cmidrule{2-12}
         {} & {Prior} & {55.71} & {28.48} & {39.60} & {53.24} & {50.27} & {48.87} & {\textbf{50.99}} & {35.59} & {42.05} & {44.98} \\ 
         {} & {FCR} & \textbf{{62.86}} & {\textbf{49.01}} & {\textbf{55.47}} & {\textbf{61.61}} & {\textbf{64.05}} & {\textbf{65.68}} & {50.68} & {\textbf{42.11}} & {\textbf{48.59}} & {\textbf{55.56}} \\
         \midrule
         \multirow{3.5}{*}{$\mathbb{D}_{low_{b}}$} & {\textcolor{gray}{\#Size}} & {\textcolor{gray}{58.20}} & {\textcolor{gray}{1.20}} & {\textcolor{gray}{6.20}} & {\textcolor{gray}{101.40}} & {\textcolor{gray}{3.20}} & {\textcolor{gray}{2.20}} & {\textcolor{gray}{146.80}} & {\textcolor{gray}{38.00}} & {\textcolor{gray}{2.80}} & {\textcolor{gray}{40.00}} \\
         \cmidrule{2-12}
         {} & {Prior} & {36.08} & {-} & {-} & {33.14} & {-} & {-} & {35.97} & {17.37} & {-} & {30.64} \\ 
         {} & {FCR} & {\textbf{46.39}} & {-} & {-} & {\textbf{52.47}} & {-} & {-} & {\textbf{40.87}} & {\textbf{34.74}} & {-} & {\textbf{43.62}} \\
         \bottomrule 
    \end{tabular}%
    }
\end{table*}

\textbf{Effect of different sample size $t$.} The Prior accuracy is a key metric reflecting the effectiveness of division, where lower $\mathcal{CS}$ is expected to correlate with lower Prior accuracy. Consequently, we conducted experiment to observe the impact of varying $t$ from 3 to 20 on Prior accuracy. As illustrated in Figure~\ref{fig:scNumDivide_average_acc}, there is a clear distinction in Prior accuracy on different subsets, and only a minimal number of inferences are required to reach an oscillatory state, which supports the reasonability behind basing $t$ on $|\mathbf{C}_i|$. Additionally, the number of different subsets size after division is also a crucial metric, as it directly impacts the overall cost of DCR. Therefore, Figure~\ref{fig:scNumDivide_average_size} presents the sizes of $\mathbb{D}_{other}$ and $\mathbb{D}_{low}$ across various $t$. Similar to Prior accuracy, the sizes of different subsets also stabilize in a fluctuating range with only minimal inferences.

\textbf{Effect of different dividing threshold $\mu$.} Based on the definition of sample size $t$ and the strategy of DCR in Section~\ref{sec:method_divide}, we divided the dataset into four discrete subsets according to $\mathcal{CS}$ intervals: (0.8, 1] for $\mathbb{D}_{high}$, (0.6, 0.8] for $\mathbb{D}_{med}$, (0.4, 0.6] for $\mathbb{D}_{low_{t}}$, and [0, 0.4] for $\mathbb{D}_{low_{b}}$. Considering the model's high confidence on $\mathbb{D}_{high}$ ($\mathcal{CS}$ greater than 0.8), we only report Prior accuracy, which exceeds 85\% in majority (7 out of 9) of datasets, as shown in Table~\ref{tab:diffSubset}. This indicates that the most questions in $\mathbb{D}_{high}$ are relatively simple and require no further process. Contrastingly, $\mathbb{D}_{med}$ demonstrate moderate Prior accuracy and achieve improvements via FCR in minority (2 out of 9) datasets. In fact, the $\mathcal{CS}$ for each item in $\mathbb{D}_{med}$ belongs to (0.6, 0.8], indicating that despite the model generates diverse answers, it predominantly focuses on a specific one. This introduces a significant challenge to enhance LLM's performance by correcting its previously generated mistakes, rendering the gains through FCR as limited. In addition, referring to original DCR, $\mathbb{D}_{low_{t}}$ and $\mathbb{D}_{low_{b}}$ comes from further dividing of the $\mathbb{D}_{low}$, where the former has higher $\mathcal{CS}$. Therefore, the average Prior accuracy of $\mathbb{D}_{low_{b}}$ is only 30.64\%, markedly below 44.98\% of $\mathbb{D}_{low_{t}}$, and significantly inferior to others. Meanwhile, through the conquer phase in DCR, we achieve an average accuracy improvement of 10.58\% and 12.98\% for $\mathbb{D}_{low_{t}}$ and $\mathbb{D}_{low_{b}}$, respectively. However, more than half of the $\mathbb{D}_{low_{b}}$ across various datasets contain a minimal number of data items, making it difficult to reliably report accuracy or effectively improve performance for entire dataset. Therefore, we instituted the threshold $\mu$ as 0.6 to conduct dataset dividing.

\begin{wrapfigure}{r}{.45\textwidth}
    \vspace{-1mm}
    \centering    
    \includegraphics[width=.45\textwidth]{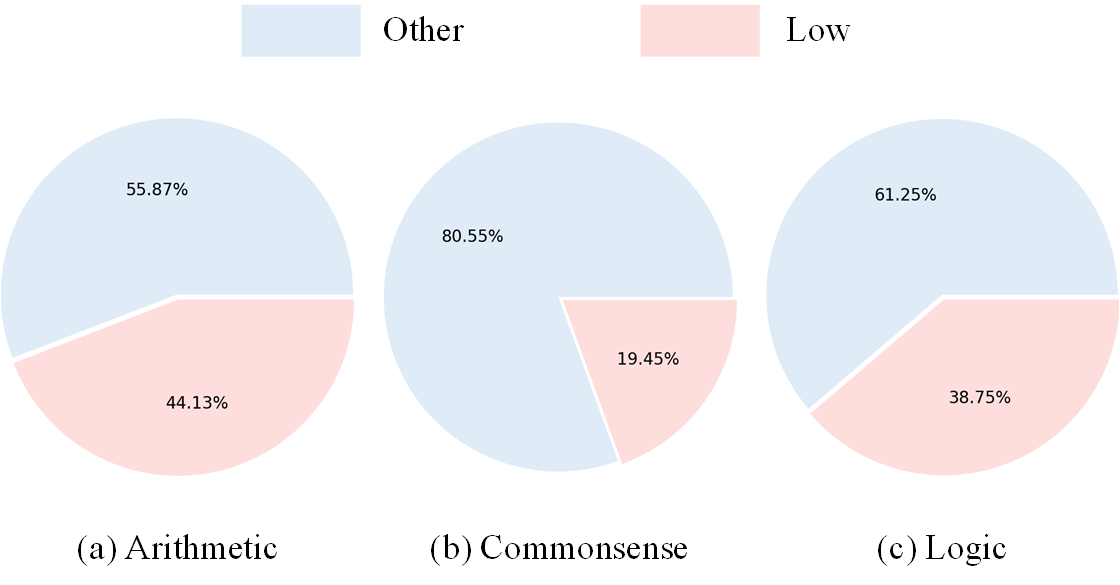}
    \caption{Distribution of different subsets among three tasks. See Figure~\ref{fig:prosubsetDiffDataset} for details about each dataset.}
    \label{fig:prosubsetDiffTask}
    \vspace{-5mm}
\end{wrapfigure}
\textbf{The distribution of different subsets.} Incorporating the dividing results, we conducted a visual statistical analysis to examine the distribution of different confidence subsets among three reasoning tasks, as shown in Figure~\ref{fig:prosubsetDiffTask}. The proportion of other subsets all exceeds 50\% in different tasks and even surpasses 80\% in commonsense class, which means that we can achieve high accuracy on a substantial portion of data without complex processing. Therefore, based on DCR, we can concentrate more resources on low confidence subsets while effectively avoid redundant process on other ones, which significantly reduce overall expenditure.

\subsubsection{Study for conquer stage} \label{sec:study_for_conquer}
\begin{table}[t]
    \centering
    \caption{Comparison of problem-solving accuracy (\%) for conquering different subsets.}
    \label{tab:conquer_diffset}
    \resizebox{\textwidth}{!}{%
    \begin{tabular}{c|ccccccccc|cc}
         \toprule
         \multirow{2.5}{*}{\makecell[c]{Conquer\\Subset}} & \multicolumn{3}{c}{Arithmetic} & \multicolumn{3}{c}{Commonsense} & \multicolumn{3}{c|}{Logic} & \multirow{2.5}{*}{Avg.} & \multirow{2.5}{*}{\#Call} \\
         \cmidrule(r{1mm}){2-4} \cmidrule(lr{1mm}){5-7} \cmidrule(l{1mm}){8-10}
         {} & {AQ.} & {Alg.} & {Math.} & {CMS.} & {OB.} & {ARC.} & {Rid.} & {Logi.} & {Rec.} & {} \\
         \midrule
         {$\mathbb{D}_{med}\&\mathbb{D}_{low}$} & {69.92} & {45.40} & {\textbf{67.04}} & {77.36} & {86.16} & {89.55} & {67.40} & {48.40} & {\textbf{62.36}} & {68.18} & {6.78} \\
         {$\mathbb{D}_{low}$} & {\textbf{71.02}} & {\textbf{48.60}} & {66.52} & {\textbf{77.97}} & {\textbf{86.80}} & {\textbf{89.79}} & {\textbf{68.81}} & {\textbf{50.27}} & {61.96} & {\textbf{69.08}} & {\textbf{5.79}} \\
         \bottomrule 
    \end{tabular}%
    }
\end{table}

\textbf{Different conquer subsets.}
Building upon Section~\ref{sec:study_for_divide}, we retain $\mathbb{D}_{med}$ and combine $\mathbb{D}_{low_{t}}$ and $\mathbb{D}_{low_{b}}$ into $\mathbb{D}_{low}$ to compare the impact of conquering different subsets, as shown in Table~\ref{tab:conquer_diffset}. $\mathbb{D}_{low}$, being a smaller subset, requires an average sample size of 5.79, which is 0.99 lower than conquering $\mathbb{D}_{med}$ and $\mathbb{D}_{low}$ together. Furthermore, according to Table~\ref{tab:diffSubset}, additional interventions on $\mathbb{D}_{med}$ by FCR yield marginal benefits. So conquering $\mathbb{D}_{med}$ and $\mathbb{D}_{low}$ together results in enhanced accuracy for only two datasets, which steers us to pay more attention solely on $\mathbb{D}_{low}$ in conquer stage.

\begin{wraptable}{r}{.5\textwidth}
    \vspace{-3mm}
    \centering
    \caption{Accuracy (\%) with different reasoning methods on $\mathbb{D}_{low}$ in AQuA and CMSQA.}
    \label{tab:diffMethod}
    \resizebox{.5\textwidth}{!}{%
    \begin{tabular}{c|cc|c}
         \toprule
         {Method} & {AQ.} & {CMS.} & {Average} \\
         \midrule
         {ManualCoT} & {43.21} & {56.96} & {50.09} \\
         {Active-Prompt} & {42.28} & {\textbf{57.88}} & {50.08} \\
         {PHP} & {44.49} & {-} & {-} \\
         \midrule
         {Zero-Shot-CoT} & {44.46} & {45.23} & {44.85} \\
         {Role-Play Prompting} & {48.20} & {46.79} & {47.50} \\
         {FCR} & {\textbf{49.45}} & {54.39} & {\textbf{51.92}} \\
         \bottomrule 
    \end{tabular}%
    }
\end{wraptable}

\textbf{Different reasoning methods.} In this section, we compared our proposed zero-shot based FCR with some representative few-shot works: ManualCoT (i.e. CoT) \citep{manualCOT}, Active-Prompt \citep{activePrompting}, and PHP \citep{PHP}, as well as some zero-shot methods: Zero-Shot-CoT \citep{zeroShotCOT} and Role-Play Prompting \citep{role-play-prompting}. Considering diverse datasets employed by these methods, we chose $\mathbb{D}_{low}$ from two widely utilized datasets (AQuA and CMSQA) for this comparison, and evaluated performance based on a single sample size. Notably, we solely conducted PHP on AQuA since it only reported results on arithmetic tasks. As shown in Table~\ref{tab:diffMethod}, FCR achieves the highest accuracy of 49.45\% on AQuA, and demonstrates competitive performance with few-shot methods on CMSQA. This presents a similar trend in Table~\ref{tab:zt_ft_cmp} of Appendix~\ref{app:zt_ft_cmp}, where Zero-Shot-CoT approaches or even exceeds Few-Shot-CoT on multiple datasets, yet it still far behind on CMSQA. Furthermore, FCR exhibits the highest average accuracy surpassing the sub-optimal zero-shot based Role-Play Prompting by 4.42\%, which highlights the strong efficacy of our method without additional human labor.

\begin{minipage}[c]{.385\textwidth}
    \captionof{table}{Accuracy (\%) on unsolved subsets with different construction methods of choices list.}
    \label{tab:choicesNum_effect}
    \resizebox{\textwidth}{!}{%
    \begin{tabular}{c|cccc}
         \toprule
         {Setting} & {AQ.} & {CMS.} & {OB.} & {Rid.} \\
         \midrule
         {List1} & {2.16} & {0.97} & {0.57} & {1.97} \\
         {List2.1} & {21.65} & {47.58} & {31.03} & {28.11} \\
         {List2.2} & {29.00} & {62.10} & {45.98} & {\textbf{49.25}} \\
         {List3} & {18.18} & {29.46} & {32.76} & {24.04} \\
         {List4} & {26.41} & {65.01} & {\textbf{58.05}} & {44.02} \\
         {List5} & {\textbf{29.39}} & {\textbf{75.52}} & {48.85} & {45.70} \\         
         \bottomrule 
    \end{tabular}%
    }
\end{minipage}%
\hfill
\begin{minipage}[c]{0.3\textwidth}
    \includegraphics[width=\textwidth]{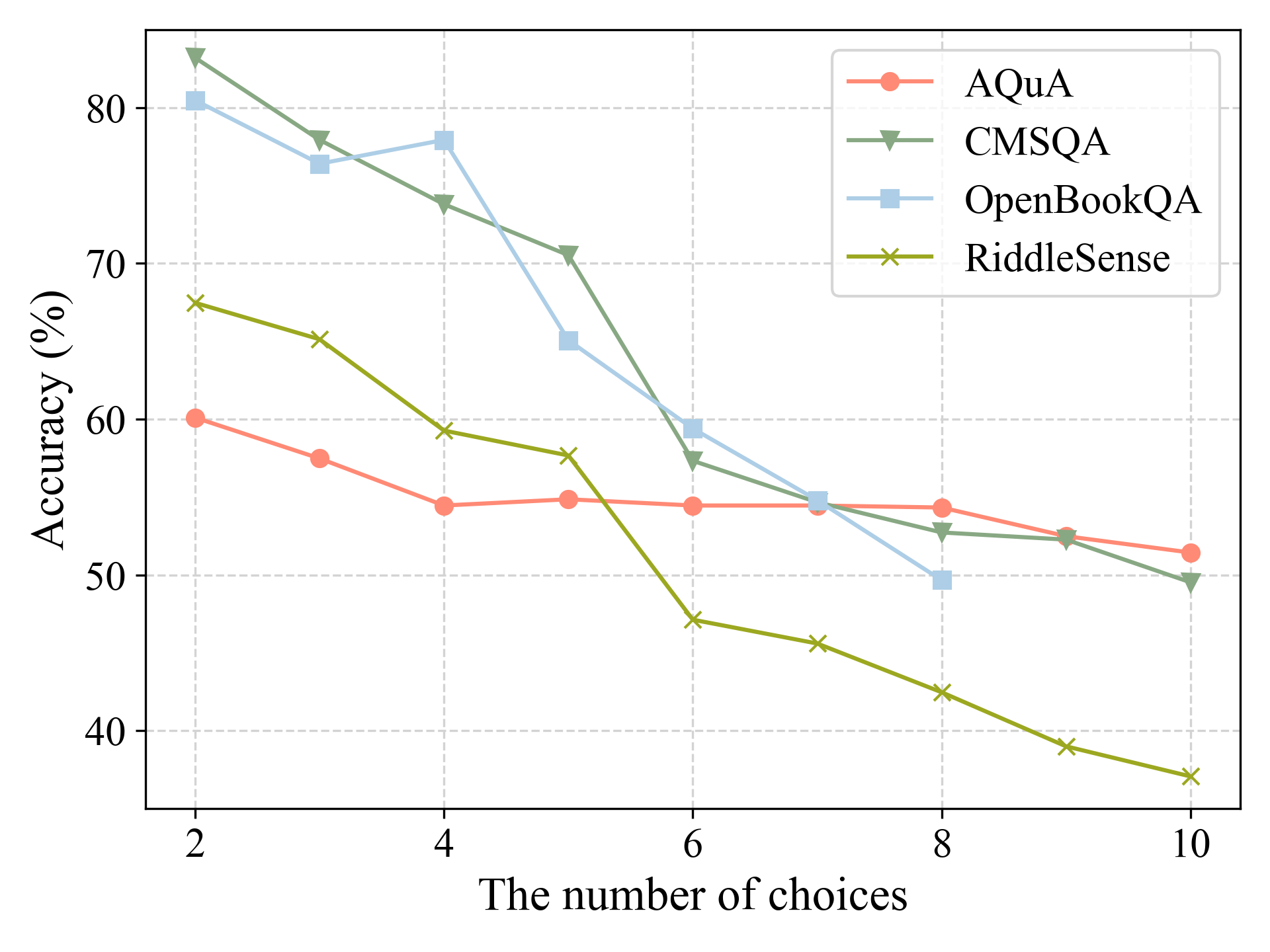}
    \captionof{figure}{Accuracy with different number of choices.}
    \label{fig:ac_choiceNum_cmp}
\end{minipage}%
\hfill
\begin{minipage}[c]{0.3\textwidth}
    \includegraphics[width=\textwidth]{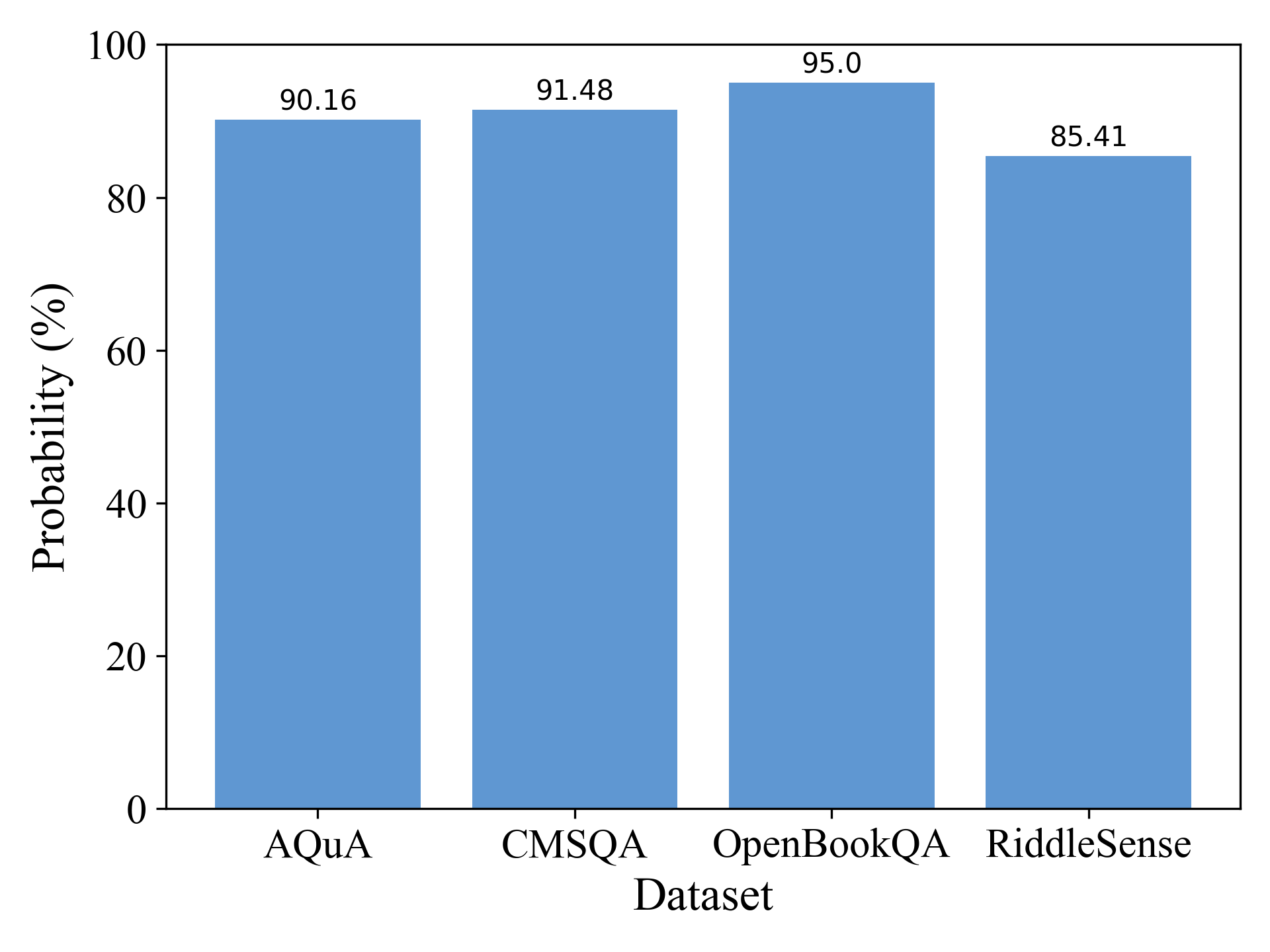}
    \captionof{figure}{Probability of correct answer in filtered list.}
    \label{fig:gold_in_SCRes}
\end{minipage}

\subsubsection{Study for irrelevant choices} \label{sec:study_irrchoices}
\textbf{Irrelevant information may distract LLM.} \citet{irrelevant} investigated the sensitivity of LLM to irrelevant information within questions and proposed to add instruction or exemplars to effectively reduce distractibility. In fact, such irrelevant information is not solely limited to the questions' context, but also contained in options list. Therefore, we conducted an analysis on accuracy with different numbers of choices, especially the impact of increasing incorrect options. As in Figure~\ref{fig:ac_choiceNum_cmp}, the accuracy exhibits a noticeable decline with more incorrect options, where we extended choices list by randomly combining wrong answers. To delve deeper, we focused on subsets of problems remaining unsolved by SC \citep{self-consistency} with 5 sample times. Then we conducted inference with various constructing methods for choices list as shown in Table~\ref{tab:choicesNum_effect}: 1) presenting the full choices list as List1; 2) combining the correct option with randomly sampled 1 or 2 incorrect ones as List2.1 or List2.2 respectively; 3) using the correct option and deduplicated results from previous five inferences as List3; 4) selecting the correct option and choices not included in earlier results as List4; 5) retaining the correct option and randomly picking one from the rest of List4 as List5. The accuracy for List1 close to 0\%, while others can significantly enhance performance. However, the correct answers for the test set are unknown in real-world scenario, which leads us to explore the feasibility of utilizing results from previous inference to filter the choices. And we quantified the probability of the correct answer in filtered choices list, as shown in Figure~\ref{fig:gold_in_SCRes}. An average 90.51\% of cases retained the correct answers, indicating that earlier results can effectively narrow down the original choices list.

\begin{wraptable}{r}{.66\textwidth}
    \centering
    \begin{minipage}[c]{.43\textwidth}
        \caption{The probability of the strong distractors appearing in the choices list.}
        \label{tab:distractors_inChoicesList}
        \resizebox{\textwidth}{!}{%
        \begin{tabular}{c|cccc|c}
             \toprule
             {Setting} & {AQ.} & {CMS.} & {OB.} & {Rid.} & {Avg.} \\
             \midrule
             {List2.1} & {51\%} & {46\%} & {26\%} & {61\%} & {46\%} \\
             {List2.2} & {22\%} & {23\%} & {12\%} & {29\%} & {21.5\%} \\
             \bottomrule 
        \end{tabular}%
        }
    \end{minipage}
    \hfill
    \begin{minipage}[c]{.21\textwidth}
        \caption{Accuracy (\%) on GSM8K.}
        \label{tab:generalization_gsm8k}
        \resizebox{\textwidth}{!}{%
        \begin{tabular}{c|cc}
             \toprule
             {} & {SC} & {DCR} \\
             \midrule
             {\#Call} & {7.00} & {6.23} \\
             {Acc.} & {84.75} & {\textbf{85.00}} \\
             \bottomrule 
        \end{tabular}%
        }
    \end{minipage}
\end{wraptable}

\textbf{Fewer choices lead to better outcomes.} Considering the varying impacts different options have on LLMs, and drawing inspiration from \citet{irrelevant}, we posit that incorrect choices previously generated by the model-called as strong distractors-exert a more profound disruptive effect. As shown in Table~\ref{tab:choicesNum_effect}, there is a significant improvement from List3 to List4, with an average increase of 22.26\% across four datasets. Furthermore, retaining two choices (List2.2) consistently surpasses those with three choices (List2.1), which can be primarily attributed to the reduced likelihood of encountering strong distractors when only two options are reserved, as shown in Table~\ref{tab:distractors_inChoicesList}. Therefore, developing more effective strategies to identify and eliminate such strongly distracting options will become a crucial direction for our future research.

\subsubsection{Application beyond MCQs} \label{sec:generalization}
In the preceding experiments, all datasets are comprised by MCQs, where the correct answer is included in the choices list. Consequently, we ventured to apply DCR to GSM8K \citep{gsm8k}, a high quality cloze-style dataset of grade school math questions. Initially, we queried the entire test set 5 times consistent with AQuA. Then we constructed choices list based on generated answers, resulting in a new dataset named GSM8K-MCQ, which is formally equivalent to MCQ. Subsequently, we divided GSM8K-MCQ with a threshold ($\mu$) of 0.6 and applied FCR for deeper conquering. From Table~\ref{tab:generalization_gsm8k}, DCR achieves accuracy of 85\% with 6.23 sample times, superior than SC with \#Call as 7, which indicates the efficacy of our strategy to datasets beyond MCQs. 

\section{Related Work}
\textbf{LLMs reasoning for MCQs.} As a problem format listing alternative answers, MCQs are prevalent in real world and have led to numerous related datasets, such as MMLU \citep{mmlu}, BIG-bench \citep{bigbench}, AGIEval \citep{agieval}, CEVAL \citep{ceval}. Simultaneously, many works have emerged in MCQs community. \citet{MCQA} explore to integrate the question and the choices list, then guide the model to select the correct option's symbol. \citet{MC-order} discover LLM's position bias, revealing that the order of choices can significantly impact model's performance. \citet{pride} find selection bias, where LLMs display a clear preference for choosing options from specific positions. Different from these works, we explore the model's sensitivity to the number of options and verifies that filtering incorrect choices can further improve performance.

\textbf{CoT prompting in LLMs reasoning.} Recently, CoT prompting methods have significantly enhanced reasoning abilities of LLMs. As the pioneer, \citet{manualCOT} generate intermediate reasoning steps before arriving at the answer by integrating rationales into few-shot examplars. Following it, \citet{self-consistency}, \citet{least-to-most}, \citet{ToT}, \citet{got}, \citet{aot}, \citet{tabcot}, \citet{resprompt}, \citet{COP}, \citet{HtT}, \citet{GV-consistency} and \citet{backward-reasoning} are dedicated to optimizing the thinking process. \citet{PAL}, \citet{PoT}, \citet{chatcot}, \citet{yamauchi2023lpml} and \citet{jie2023design} employ external tools to disentangle computation from LLMs. \citet{autoCOT}, \citet{activePrompting}, \citet{Automate-CoT}, \citet{iter-cot}, \citet{metacot} are exploring demonstrations construction in distinct manners. \citet{mekala2023echoprompt}, \citet{step-back-prompting}, \citet{SEC}, \citet{analogical-prompting} and \citet{AgentInstruct} enable models to generate examplars by themselves. \citet{xue2023rcot}, \citet{selfcheck}, \citet{self-convince}, \citet{naturalprogram} and \citet{self-verification} introduce the concept of verification into the community. In addition, \citet{irrelevant} delves into the distractibility of LLMs by irrelevant context in questions. \citet{PHP} utilize previously generated answers as hints to progressively guide the model to the correct answer. \citet{role-play-prompting} defines specific roles for the model based on particular task. However all these works process data uniformly neglecting problem-solving difficulty. Therefore, we propose DCR to LLMs reasoning, which first divides the dataset, and then selects intricate ones to deeply process by filtering irrelevant choices.

\section{Conclusion}
In this paper, we propose DCR to enhance reasoning abilities of LLMs for MCQs by dividing dataset based on $\mathcal{CS}$ and subsequently conquering items with low $\mathcal{CS}$. Evaluation results on nine datasets across three tasks prove that DCR not only minimizes unnecessary computations for simple problems but also substantially improve performance on more intricate ones. In addition, through detailed analysis, we confirmed a positive relation between $\mathcal{CS}$ and accuracy, alongside fewer choices leading to better outcomes. Nonetheless, utilizing previously generated results to filter choices fails to effectively eliminate strong distractors and computing $\mathcal{CS}$ through SC is resource-intensive. Therefore, we will develop more efficient strategies for filtering distractions and reducing the computational demand associated with datasets division in the future.

\bibliography{reference}
\bibliographystyle{colm2024_conference}

\appendix
\section{Zero-shot vs. Few-shot} \label{app:zt_ft_cmp}
\begin{table}[h]
    \centering
    \small
    \caption{Problem-solving accuracy (\%) between Zero-Shot-CoT and Few-Shot-CoT.}
    \label{tab:zt_ft_cmp}
    \begin{tabular}{c|cccc|c}
        \toprule
        {Method} & {AQuA} & {GSM8K} & {SVAMP} & {CMSQA} & {Average} \\
        \midrule
        {Zero-Shot-CoT} & {54.86($\pm$0.67)} & {79.33($\pm$0.34)} & {78.20($\pm$1.82)} & {69.67($\pm$0.77)} & {70.52} \\
        {Few-Shot-CoT} & {53.67($\pm$0.67)} & {79.67($\pm$1.18)} & {81.60($\pm$1.34)} & {77.47($\pm$0.96)} & {73.10} \\
        \bottomrule
    \end{tabular}
\end{table}

By comparing Zero-Shot-CoT \citep{zeroShotCOT} and Few-Shot-COT \citep{manualCOT} across AQuA \citep{aqua}, GSM8K \citep{gsm8k}, SVAMP \citep{svamp} and CMSQA \citep{commonsenseqa} in Table~\ref{tab:zt_ft_cmp}, models' zero-shot capabilities are gradually nearing or even surpassing their few-shot counterparts, which is align with the conclusions in recent research \citep{code-prompting, agieval}. Therefore, our work is entirely free from human intervention and circumvents exemplars construction.

\begin{table*}[h]
    \centering
    \caption{The information statistic of datasets. For CMSQA, RiddleSense, Logical Deduction and Reclor, we select their validation sets as there are no publicly available test sets or labels. GSM8K and SVAMP are used in Section~\ref{sec:generalization} and Appendix~\ref{app:zt_ft_cmp}, which are cloze-style dataset without choices list. Particularly, for Logical Deduction, there are 60 questions with $|\mathbf{C}_i|$ as 3, 100 questions as 5, and 140 questions as 7. Therefore, given that 20\% questions have 3 choices, we make a compromise and choose $t$ as 4.}
    \label{tab:dataset}
    \resizebox{\textwidth}{!}{%
    \begin{tabular}{c|c|c|c|c|c}
         \toprule
         {Dataset} & {Task Type} & {Eval. Split} & {\#Test ($n$)} & {\#ChoicesNum ($|\mathbf{C}_i|$)} & {Infer. Times ($t$)} \\
         \midrule
         {AQuA (AQ.)} & {Arithmetic} & {Test} & {254} & {5} & {5} \\
         {Abstract Algebra (Alg.)} & {Arithmetic} & {Test} & {100} & {4} & {4} \\
         {High School Mathematics (Math.)} & {Arithmetic} & {Test} & {270} & {4} & {4} \\
         {CMSQA (CMS.)} & {Commonsense} & {Validation} & {1221} & {5} & {5} \\
         {OpenBookQA (OB.)} & {Commonsense} & {Test} & {500} & {4} & {4} \\
         {ARC Challenge (ARC.)} & {Commonsense} & {Test} & {1165} & {4} & {4} \\
         {RiddleSense (Rid.)} & {Logic} & {Validation} & {1021} & {5} & {5} \\
         {Logical Deduction (Logi.)} & {Logic} & {Validation} & {300} & {3, 5 or 7} & {4} \\
         {Reclor (Rec.)} & {Logic} & {Validation} & {500} & {4} & {4} \\
         {GSM8K} & {Arithmetic} & {Test} & {1319} & {-} & {-} \\
         {SVAMP} & {Arithmetic} & {Test} & {300} & {-} & {-} \\
         \bottomrule 
    \end{tabular}%
    }
\end{table*}

\section{Different prompts for FCR} \label{app:diffPrompt}
\begin{figure*}[htbp]
    \centering
    \includegraphics[width=\textwidth]{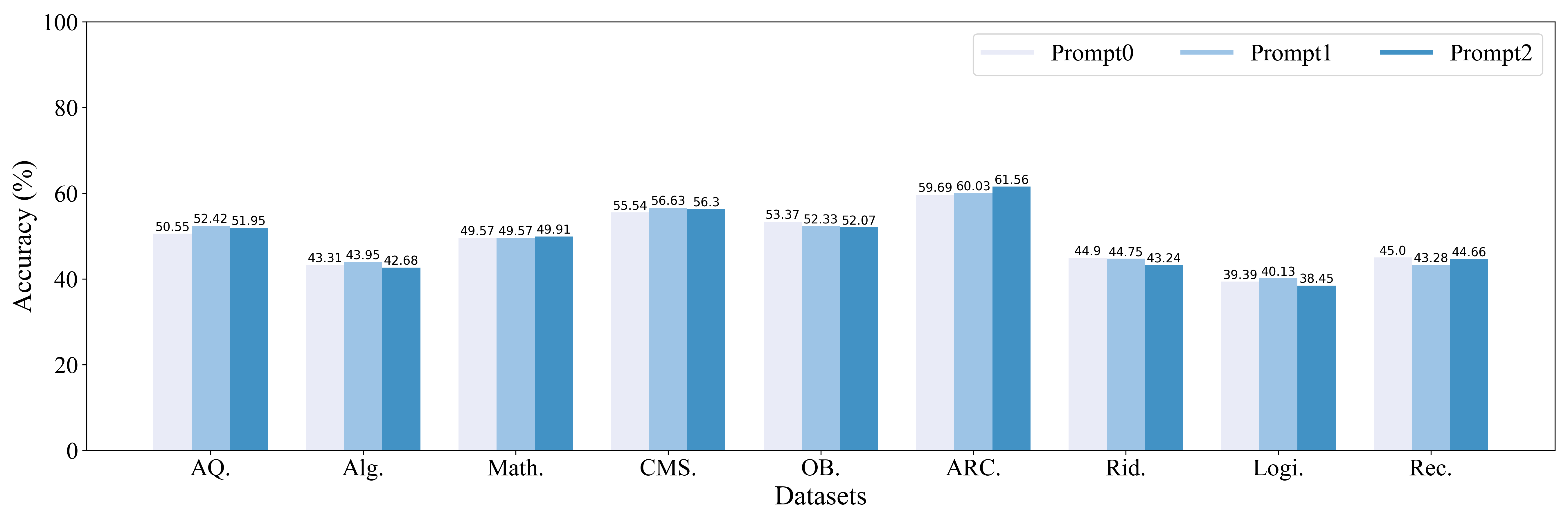}
    \caption{Accuracy of FCR on $\mathbb{D}_{low}$ for different prompts across various datasets.}
    \label{fig:diffPromptDiffDataset}
\end{figure*}

Considering the most distinctive feature of FCR is succinct choices list, we conducted a comparison using different prompts, as displayed in Figure~\ref{fig:diffPromptDiffDataset}. Specifically, ``Prompt0'' denotes ``\textcolor{gray}{Let's think step by step.}'', ``Prompt1'' is the prompt used in FCR, and ``Prompt2'' represents ``\textcolor{gray}{Let's delve deeper into this question to arrive at the best answer.}''. Across various dataset, the accuracy disparity of FCR with different prompts remains below 2\%, without a clear dominance from any single one. Therefore, we believe that the key of good performance for FCR is attributed to a briefer choices list, rather than prompt engineering.

\begin{figure*}[htbp]
    \centering
    \subfloat[AQ.]{\includegraphics[width=.33\textwidth]{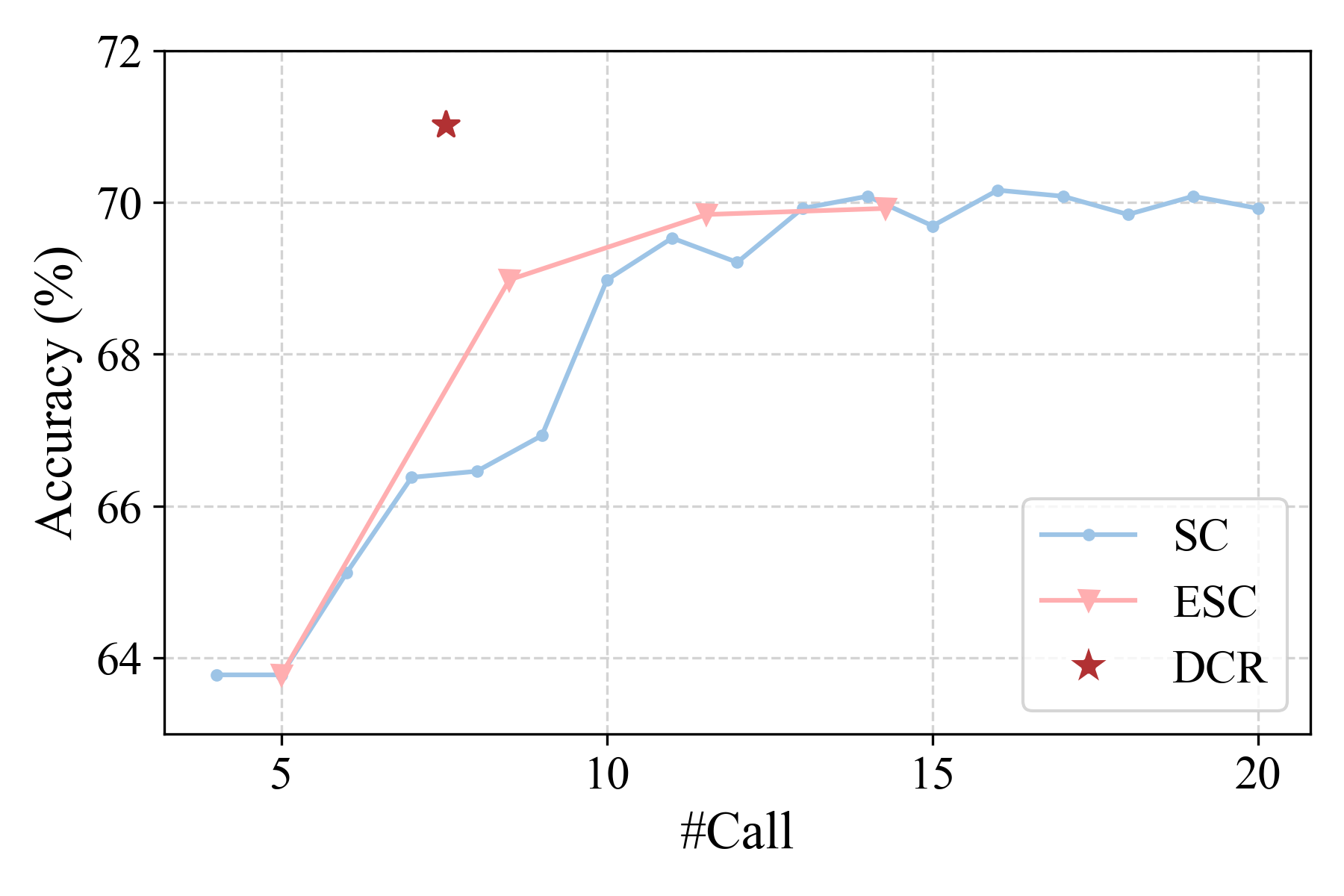}}
    \subfloat[Alg.]{\includegraphics[width=.33\textwidth]{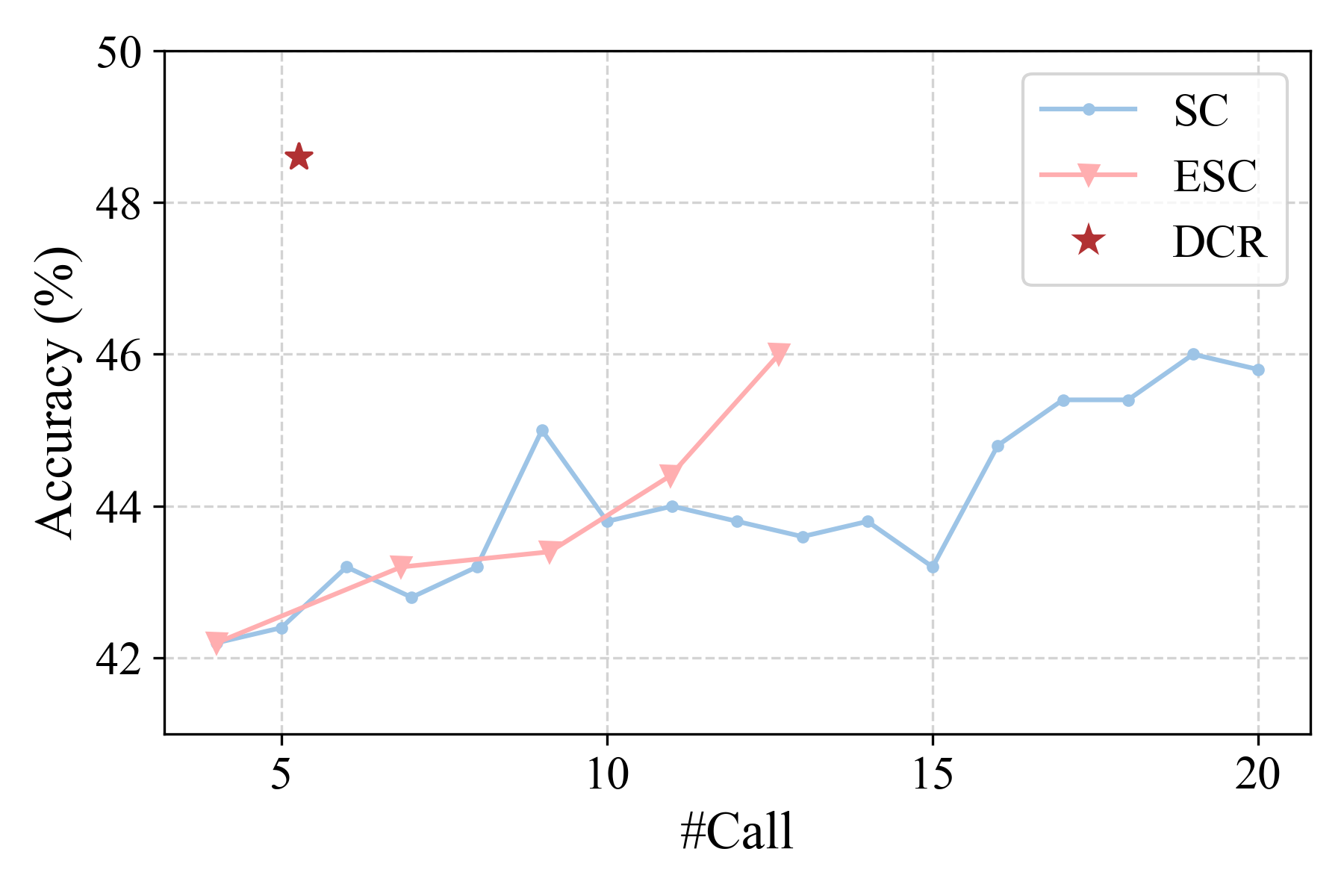}}
    \subfloat[Math.]{\includegraphics[width=.33\textwidth]{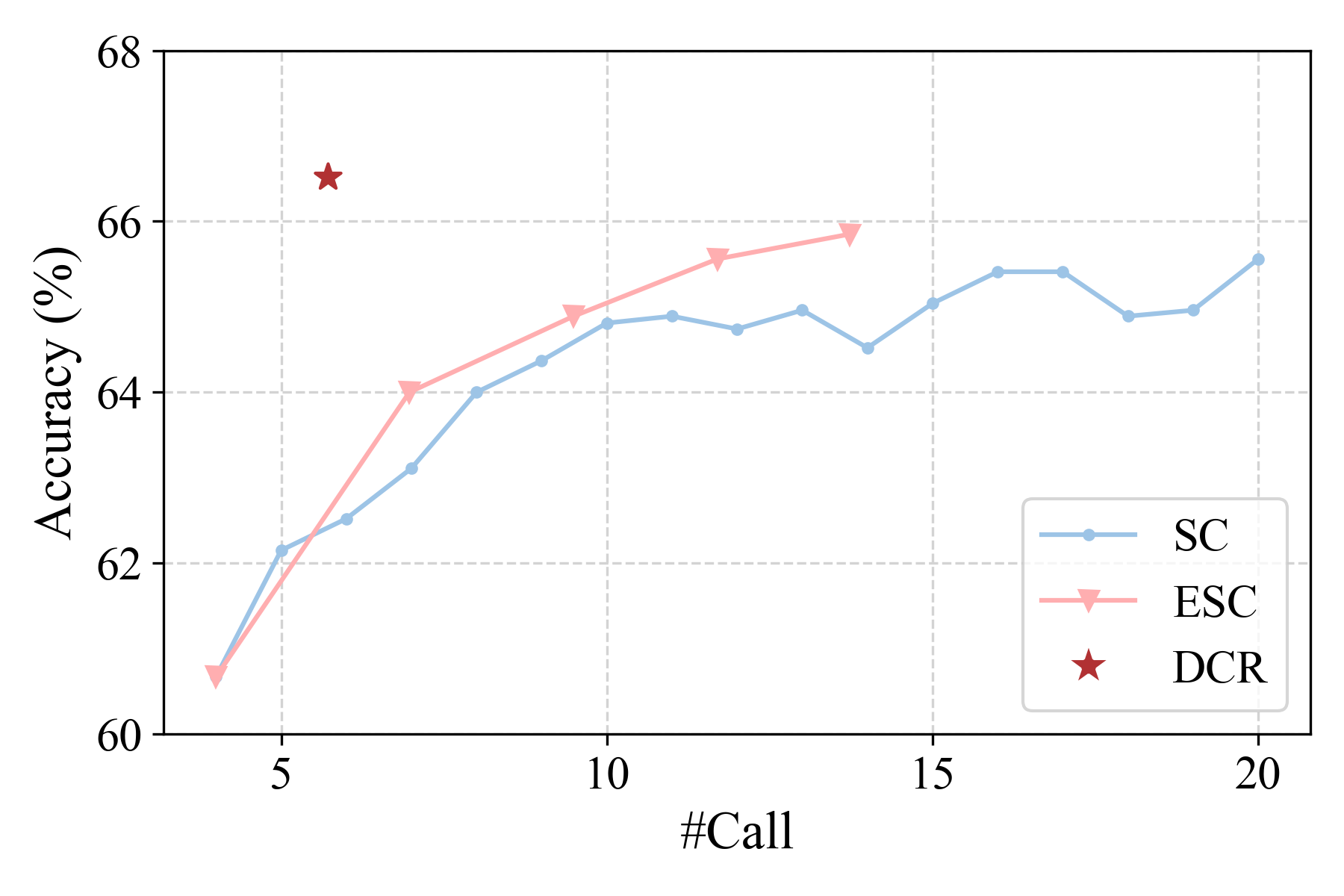}}\\
    \subfloat[CMS.]{\includegraphics[width=.33\textwidth]{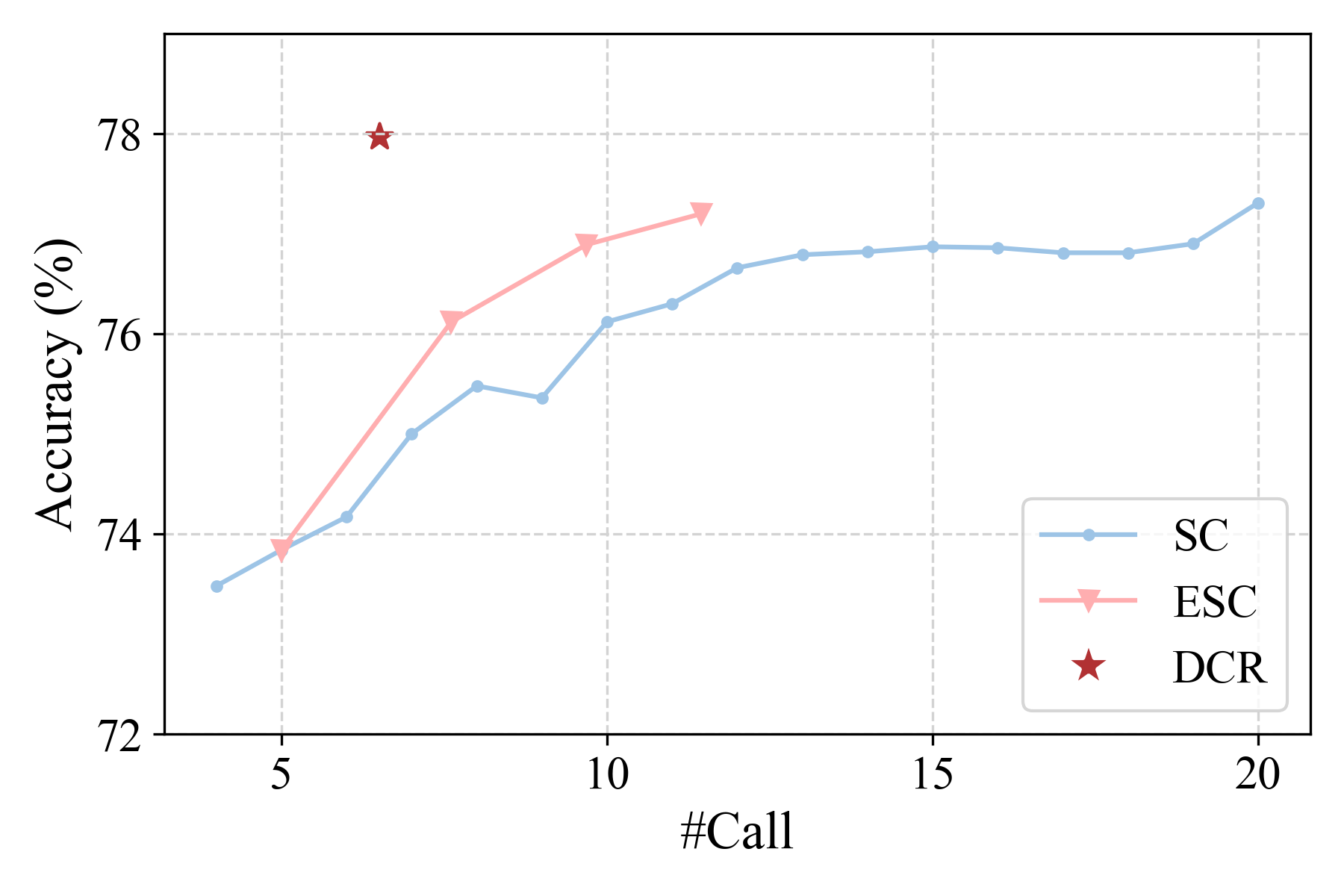}}
    \subfloat[OB.]{\includegraphics[width=.33\textwidth]{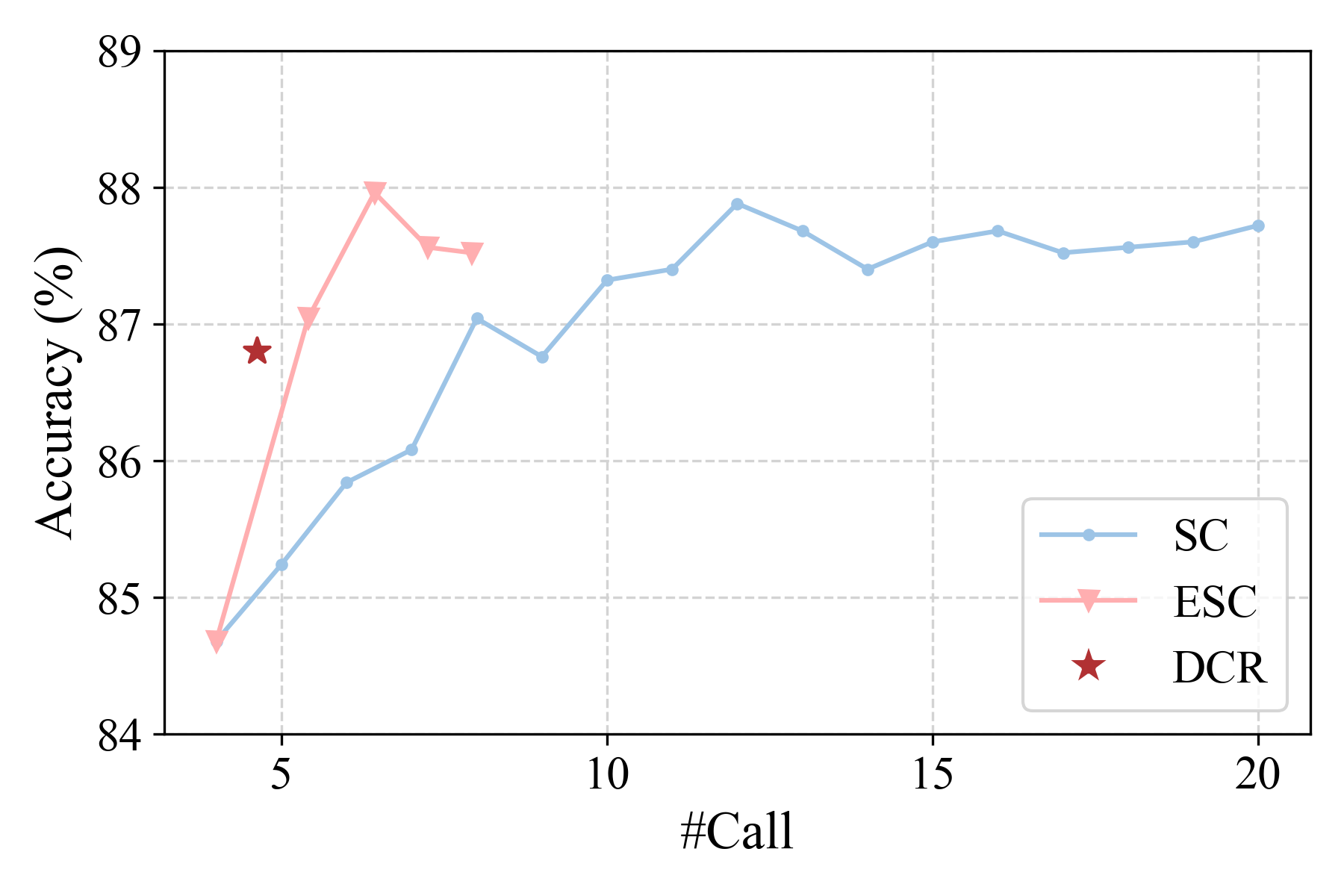}}
    \subfloat[ARC.]{\includegraphics[width=.33\textwidth]{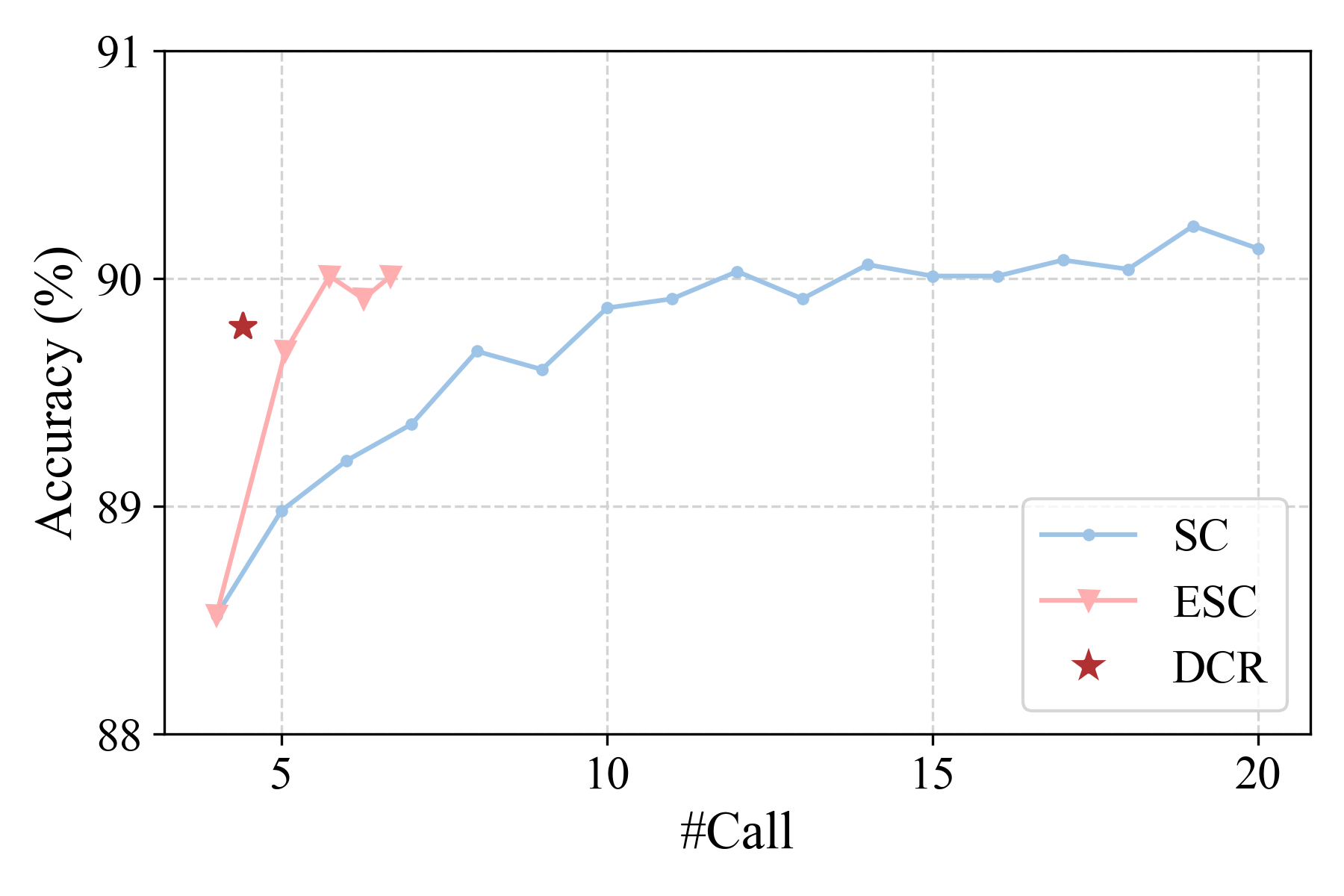}}\\
    \subfloat[Rid.]{\includegraphics[width=.33\textwidth]{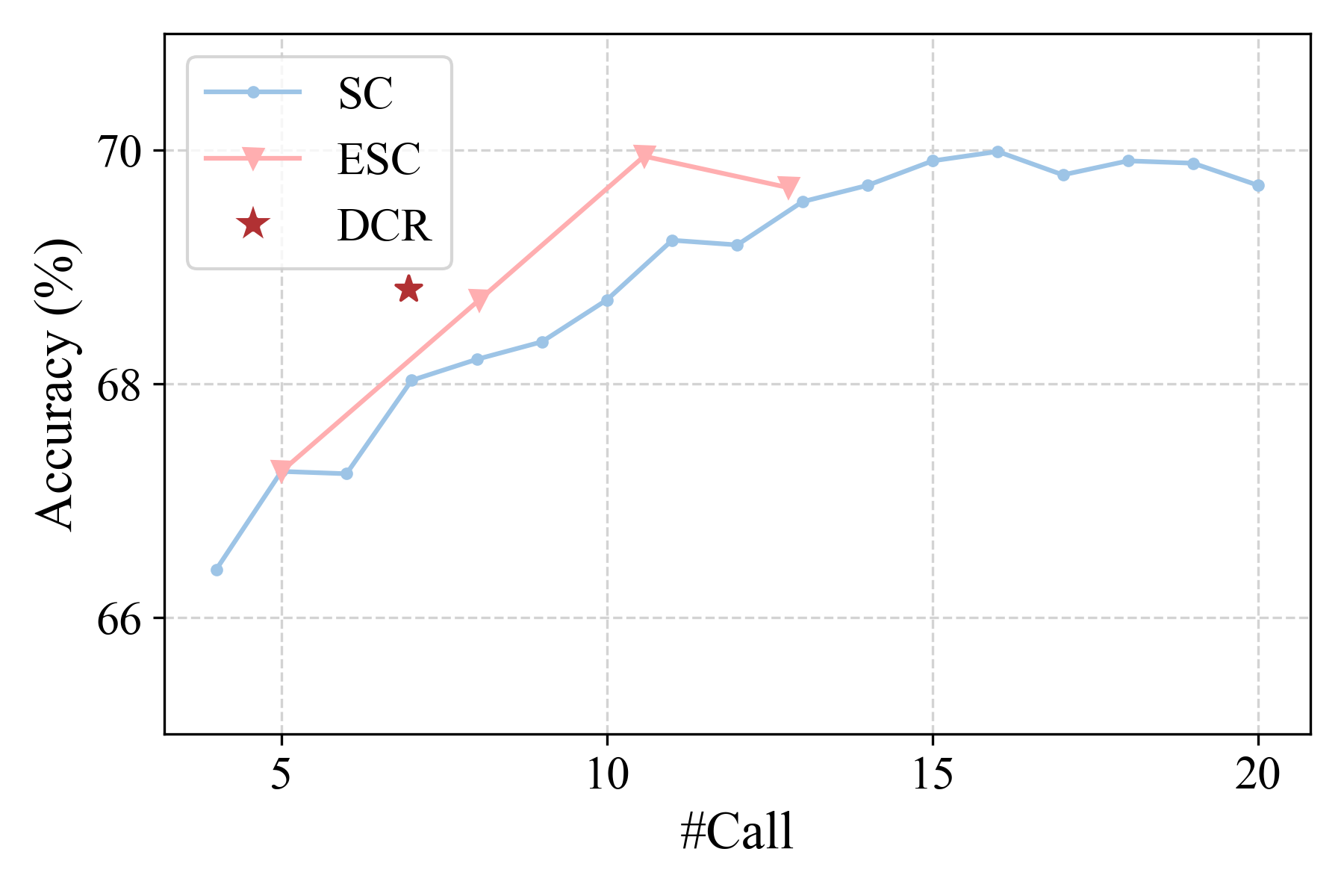}}
    \subfloat[Logi.]{\includegraphics[width=.33\textwidth]{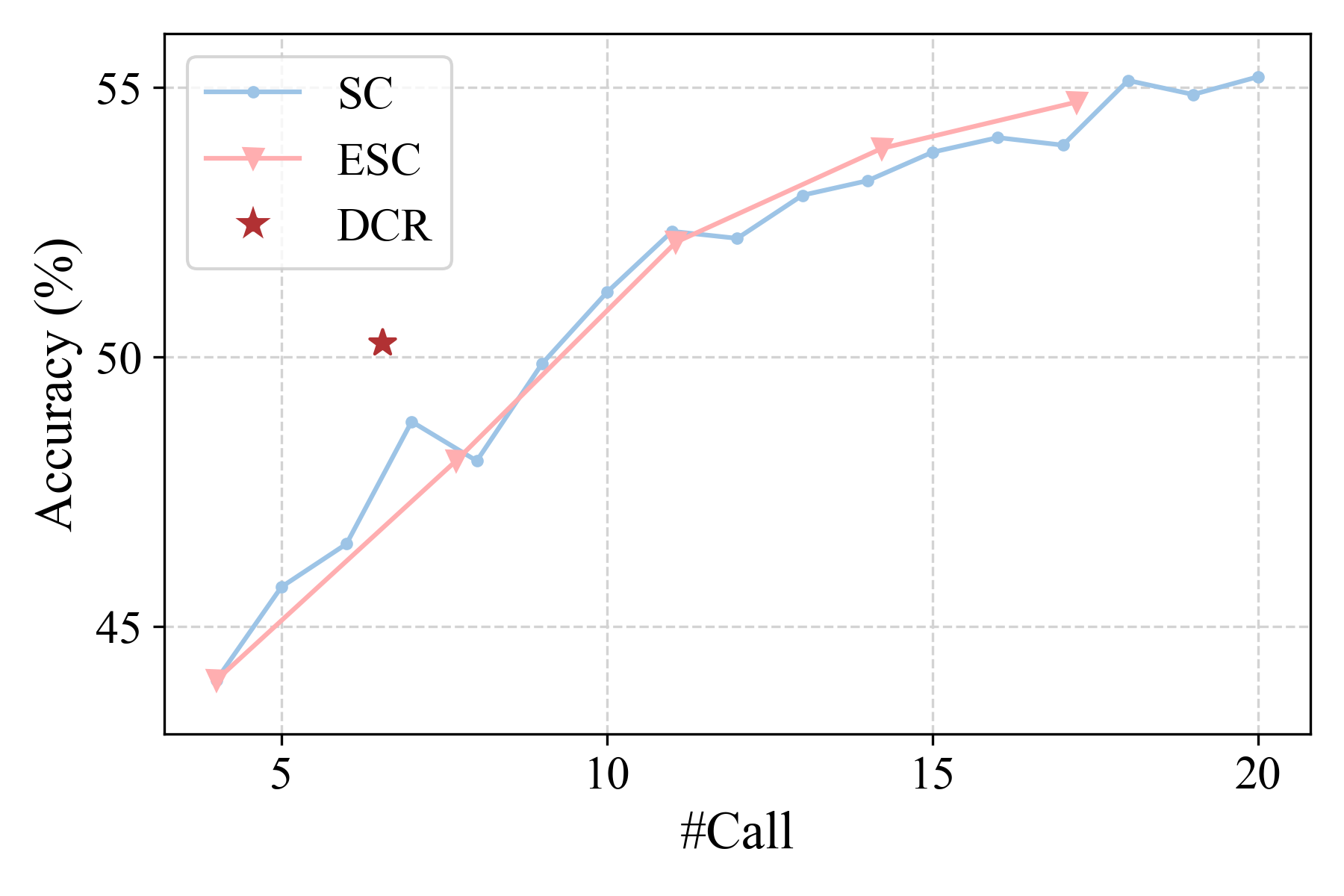}}
    \subfloat[Rec.]{\includegraphics[width=.33\textwidth]{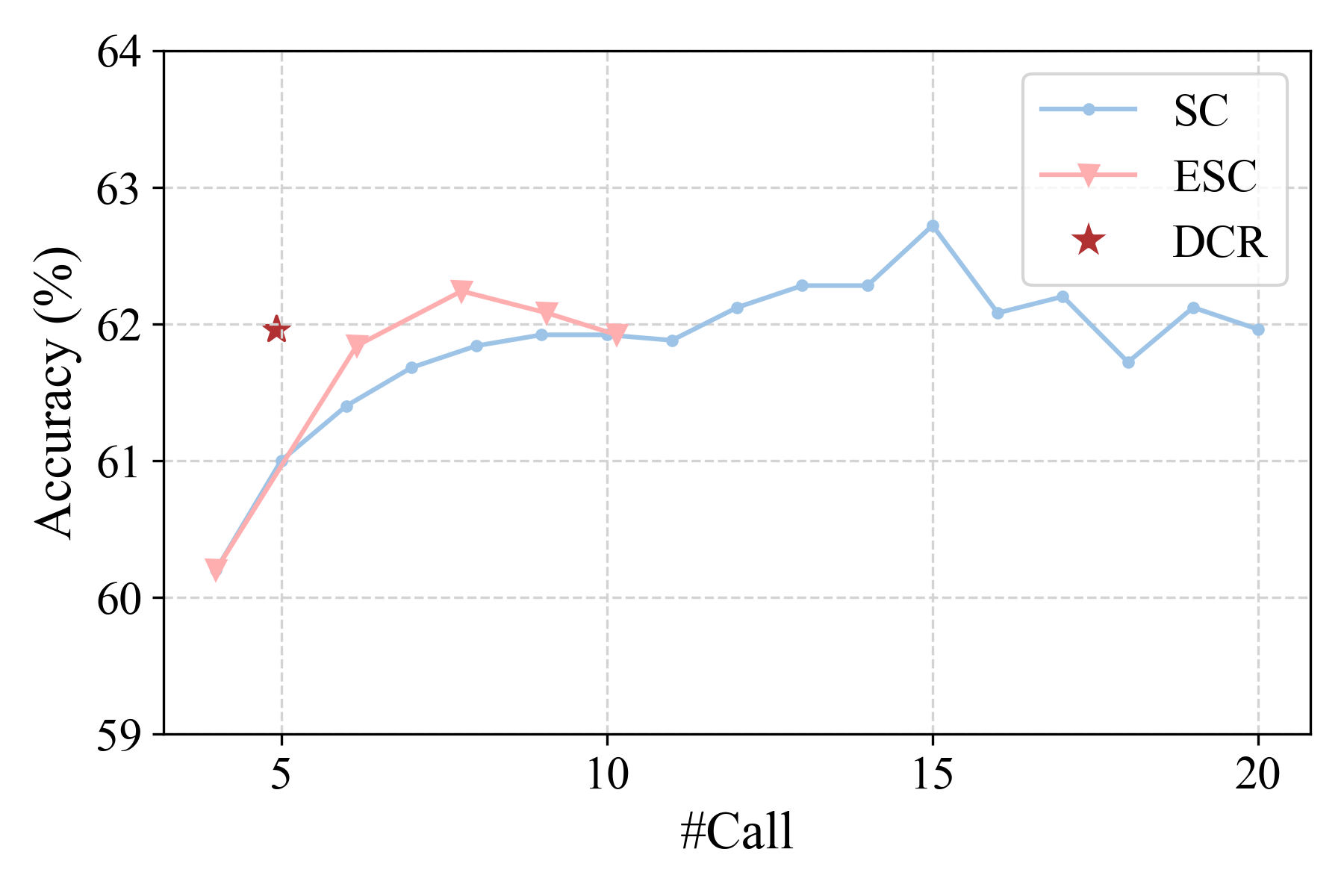}}
    \caption{Problem-solving accuracy and \#Call among different datasets.}
    \label{fig:trend_ac_cost_detail}
\end{figure*}

\begin{figure*}[htbp]
    \centering
    \includegraphics[width=\textwidth]{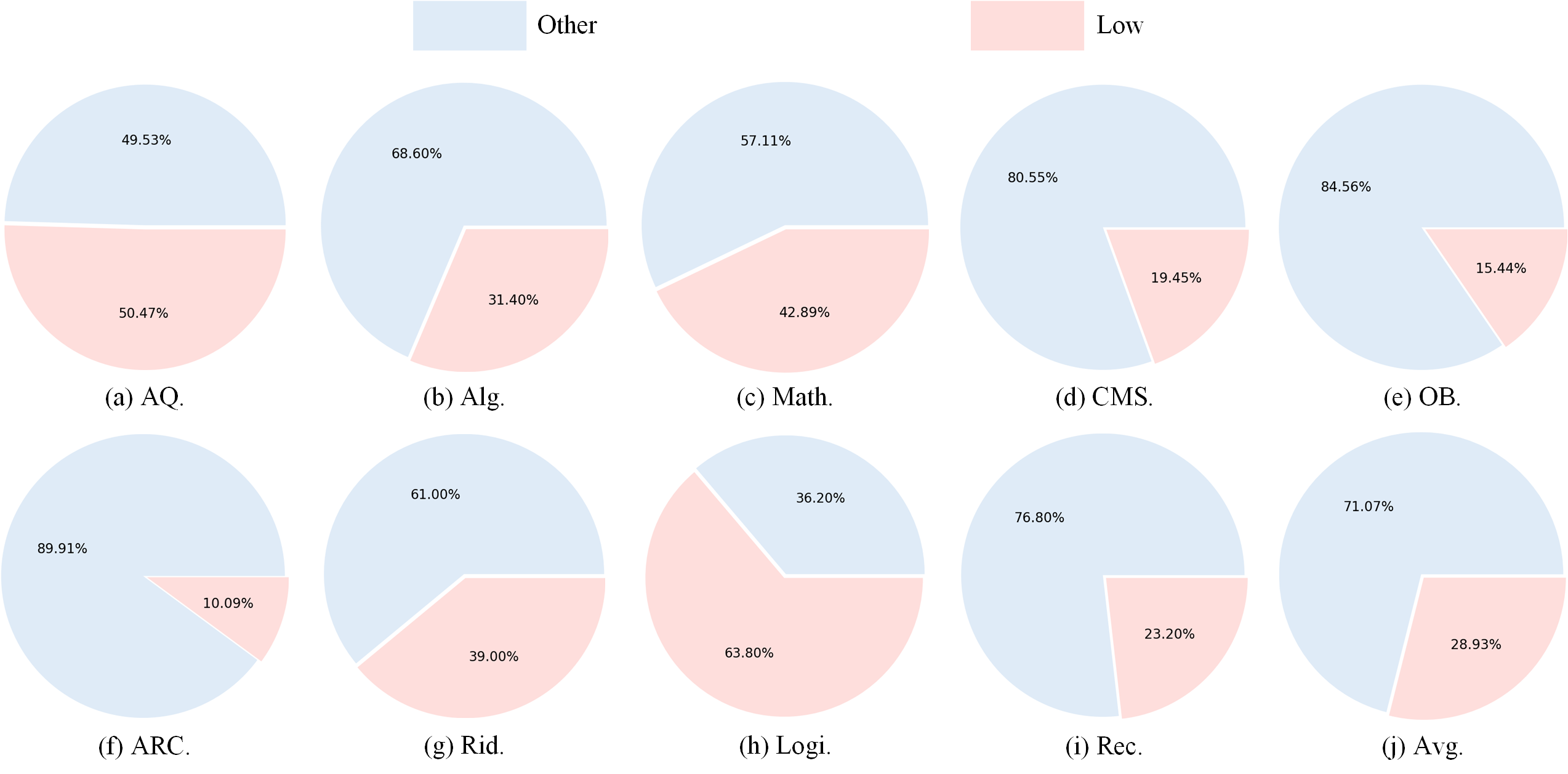}
    \caption{Distribution of different subsets among various datasets. ``Avg.'' denotes the average distribution across all datasets.}
    \label{fig:prosubsetDiffDataset}
\end{figure*}

\begin{figure*}[htbp]
    \centering
    \subfloat[AQ.]{\includegraphics[width=.33\textwidth]{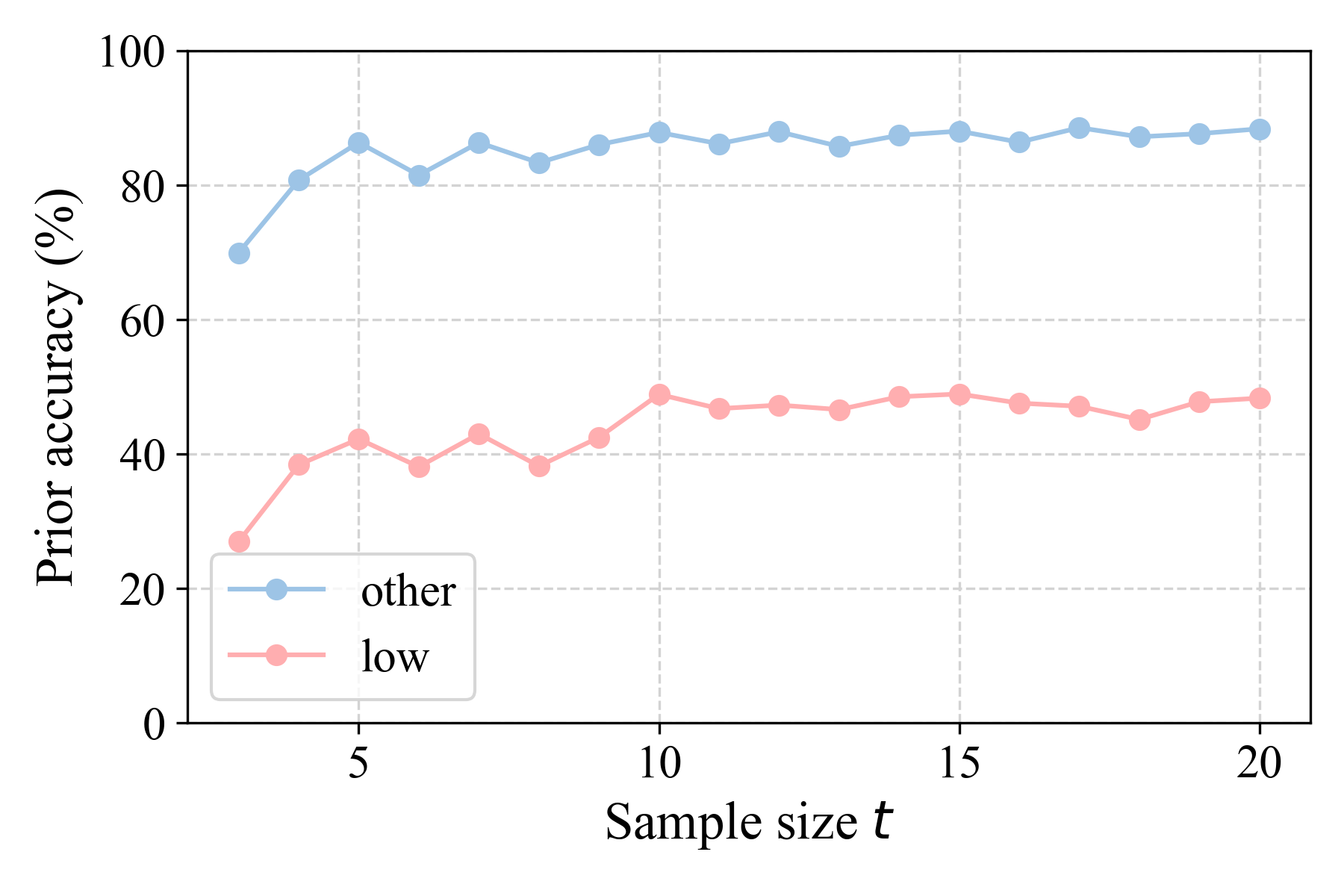}}
    \subfloat[Alg.]{\includegraphics[width=.33\textwidth]{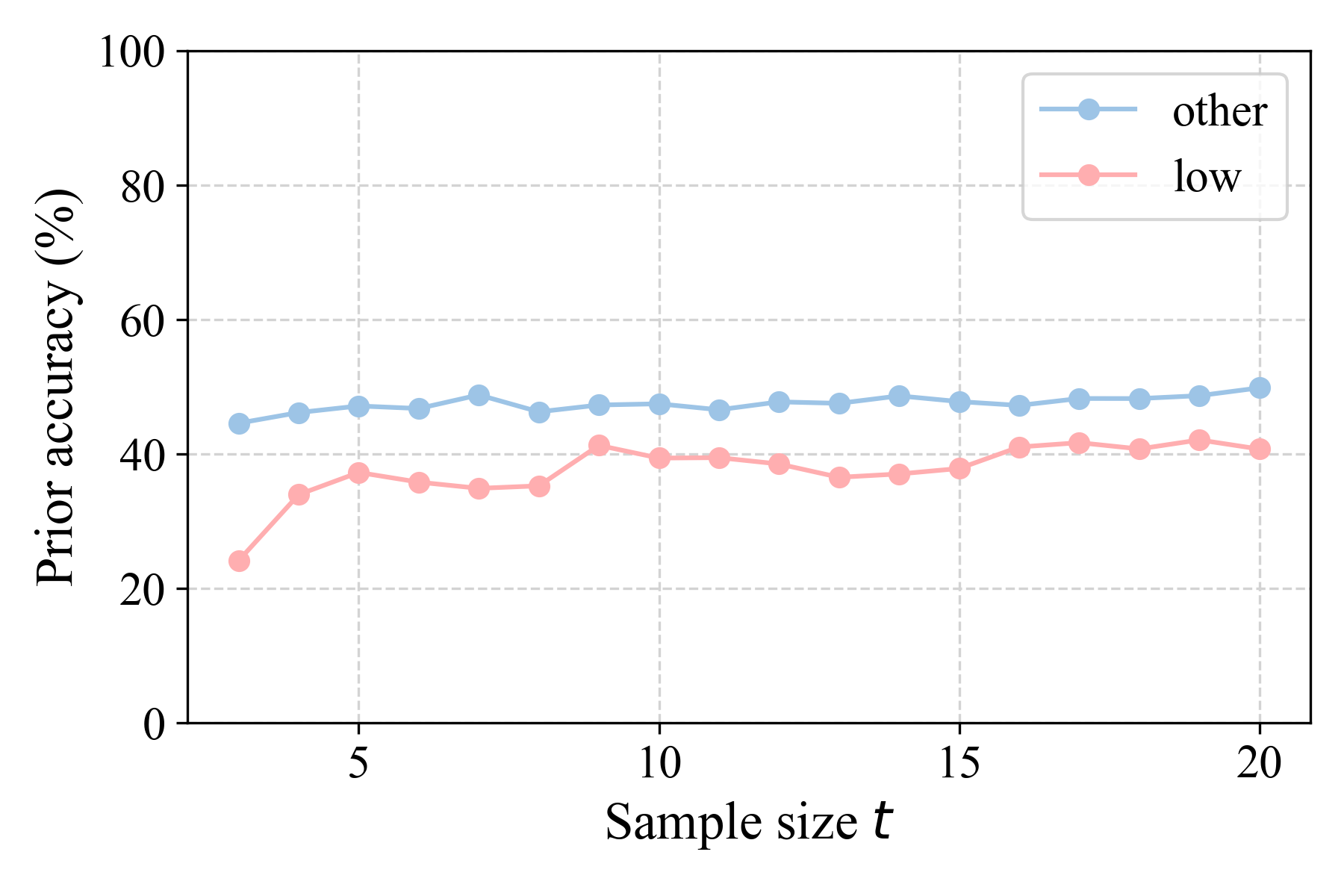}}
    \subfloat[Math.]{\includegraphics[width=.33\textwidth]{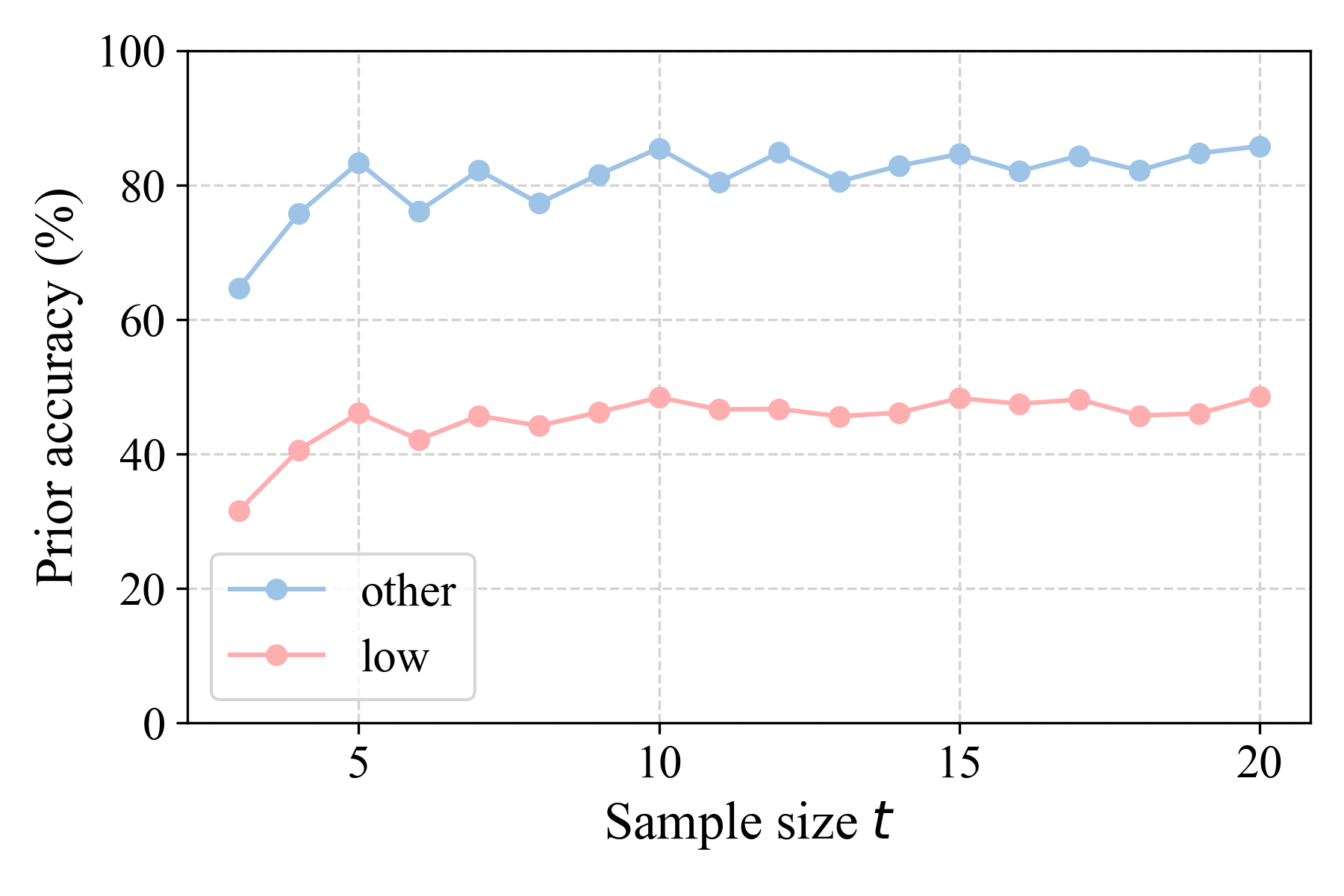}}\\
    \subfloat[CMS.]{\includegraphics[width=.33\textwidth]{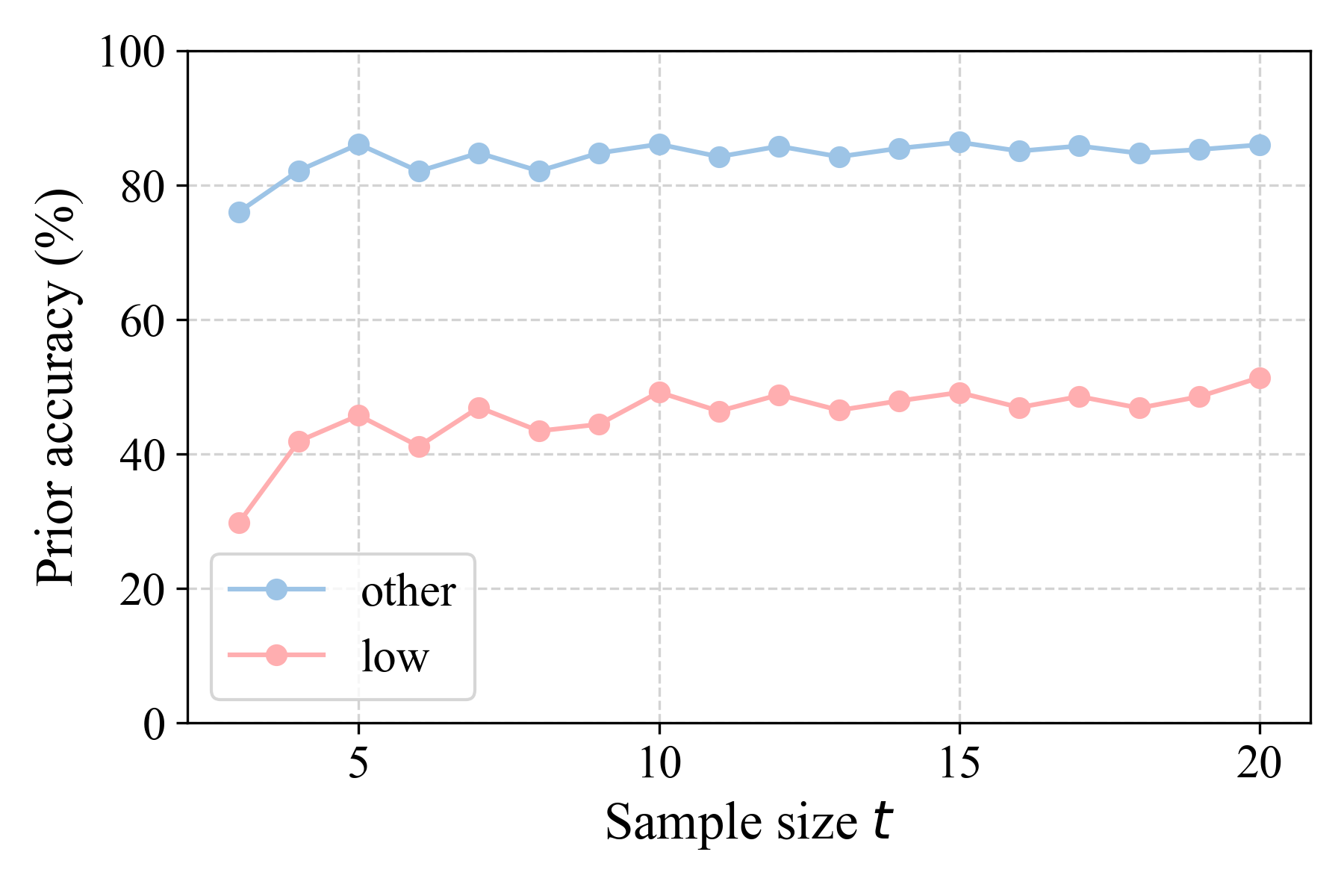}}
    \subfloat[OB.]{\includegraphics[width=.33\textwidth]{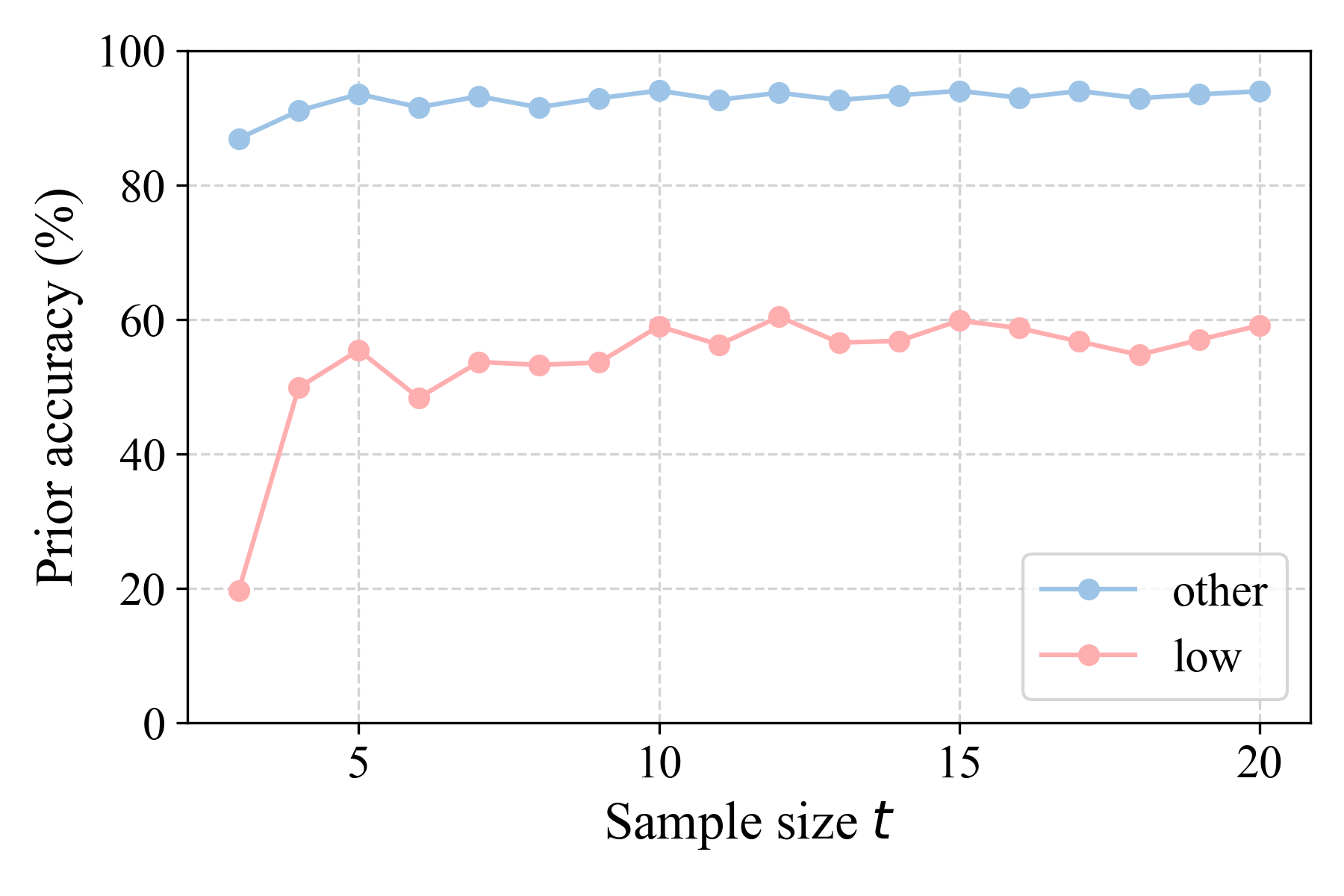}}
    \subfloat[ARC.]{\includegraphics[width=.33\textwidth]{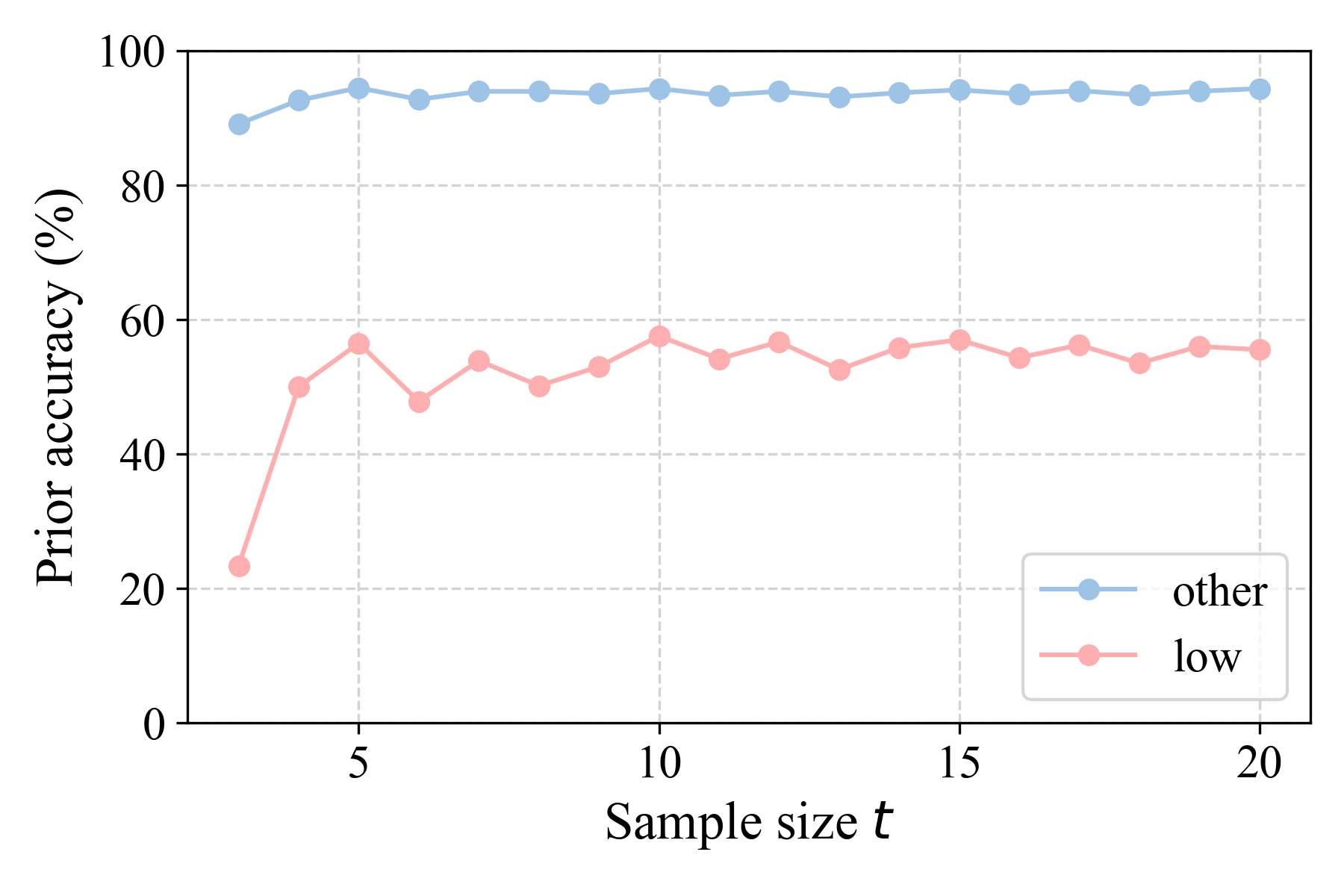}}\\
    \subfloat[Rid.]{\includegraphics[width=.33\textwidth]{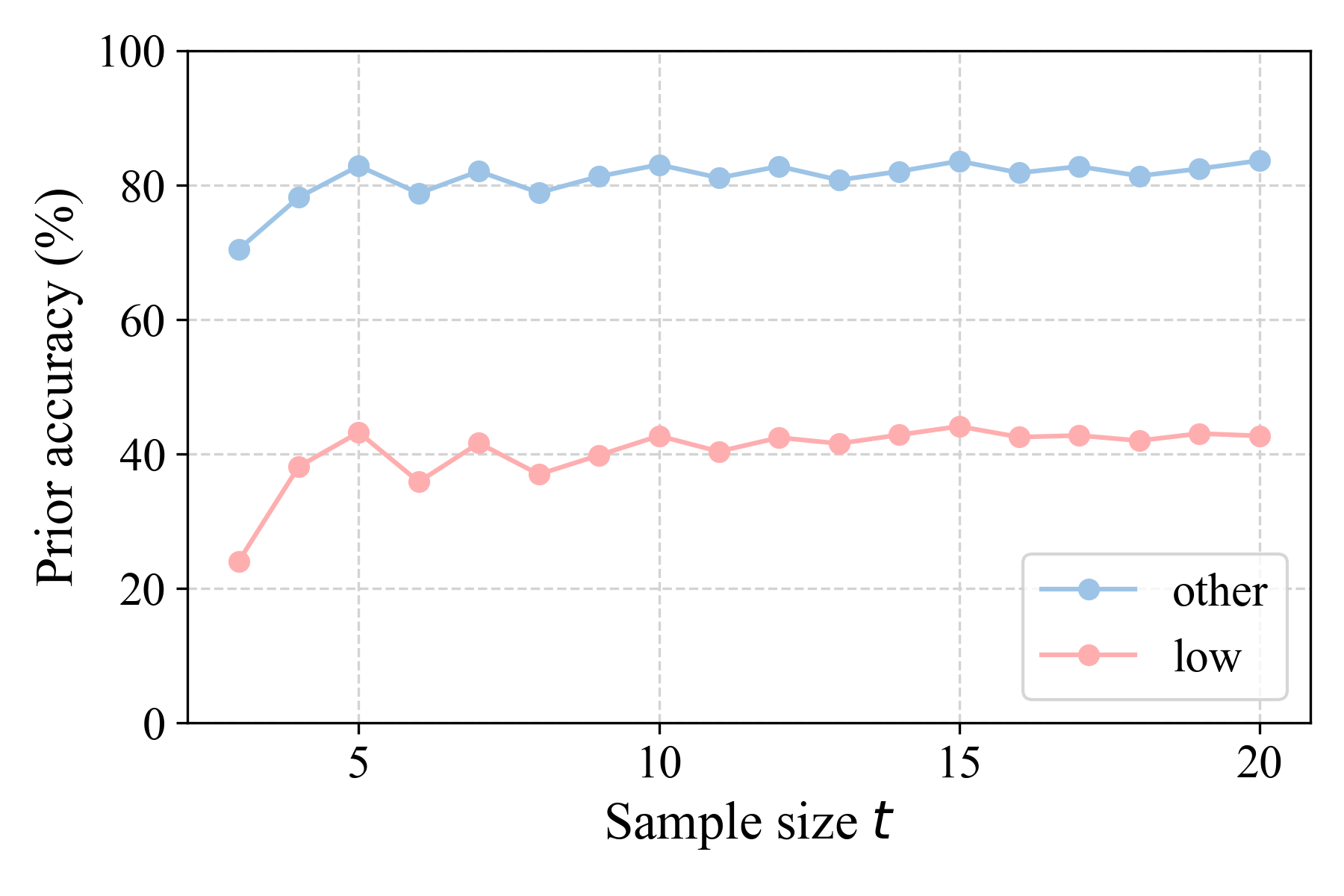}}
    \subfloat[Logi.]{\includegraphics[width=.33\textwidth]{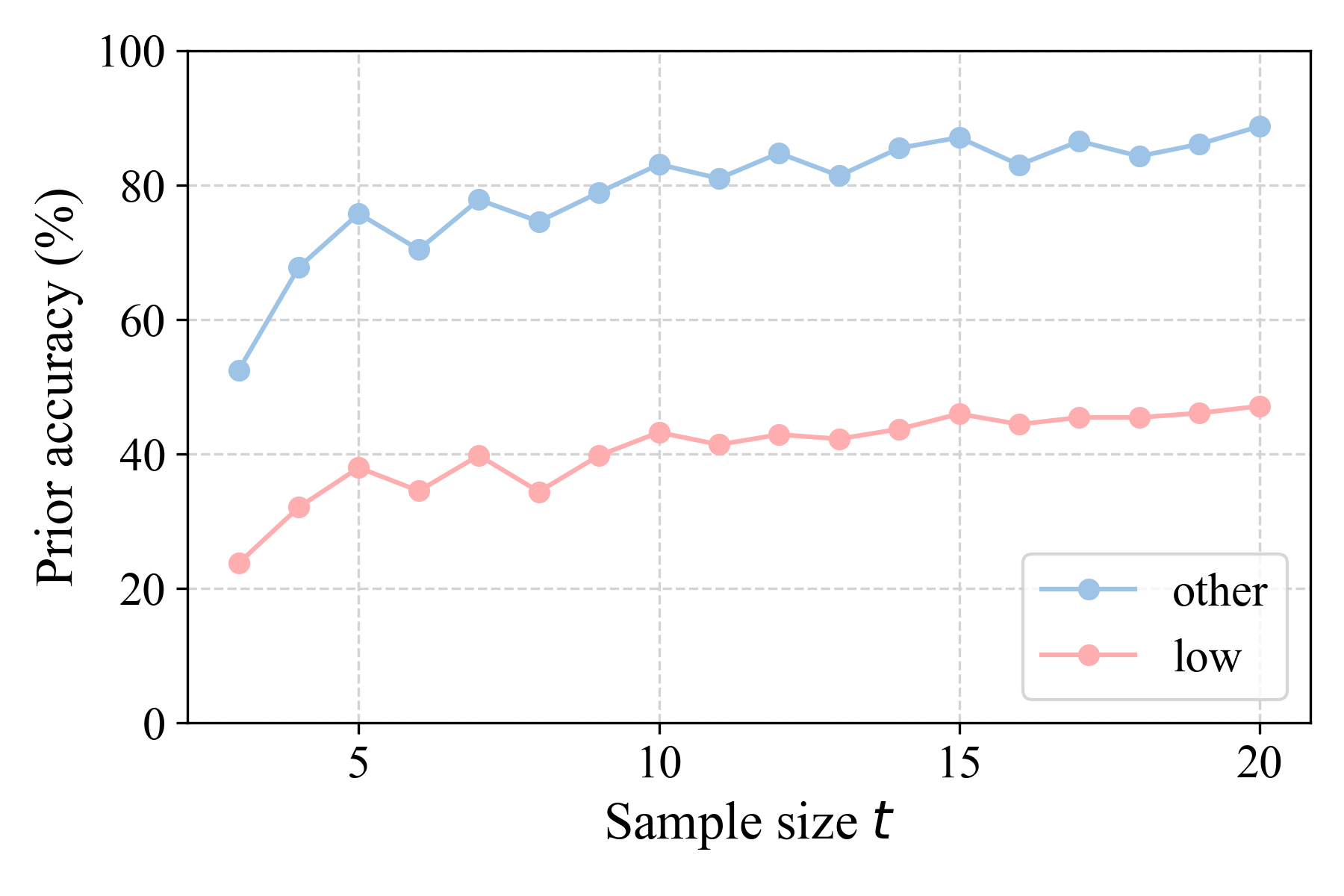}}
    \subfloat[Rec.]{\includegraphics[width=.33\textwidth]{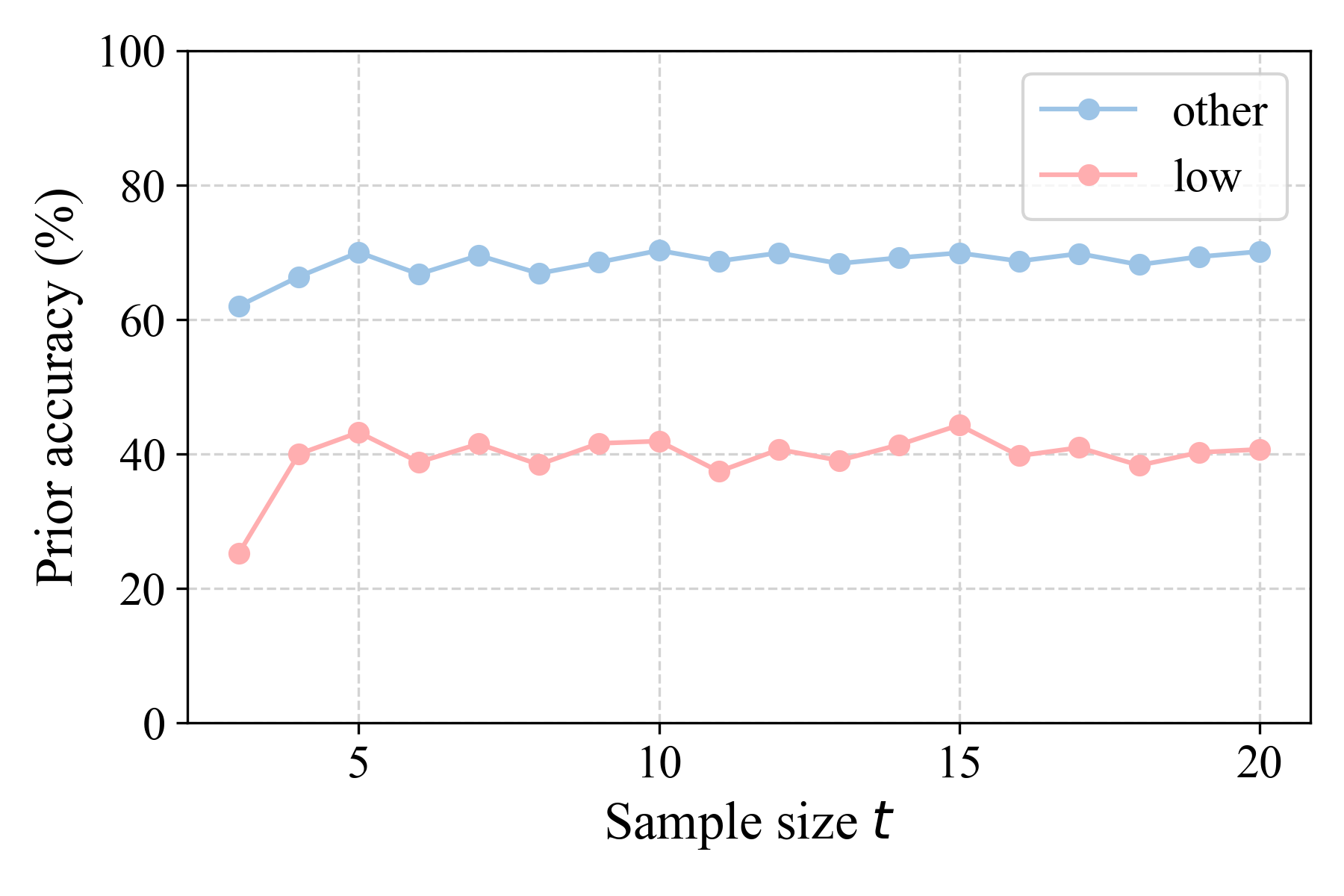}}
    \caption{Prior accuracy of different subsets for various sample size $t$ among each dataset.}
    \label{fig:scNumDivide_detail_acc}
\end{figure*}

\begin{figure*}[htbp]
    \centering
    \subfloat[AQ.]{\includegraphics[width=.33\textwidth]{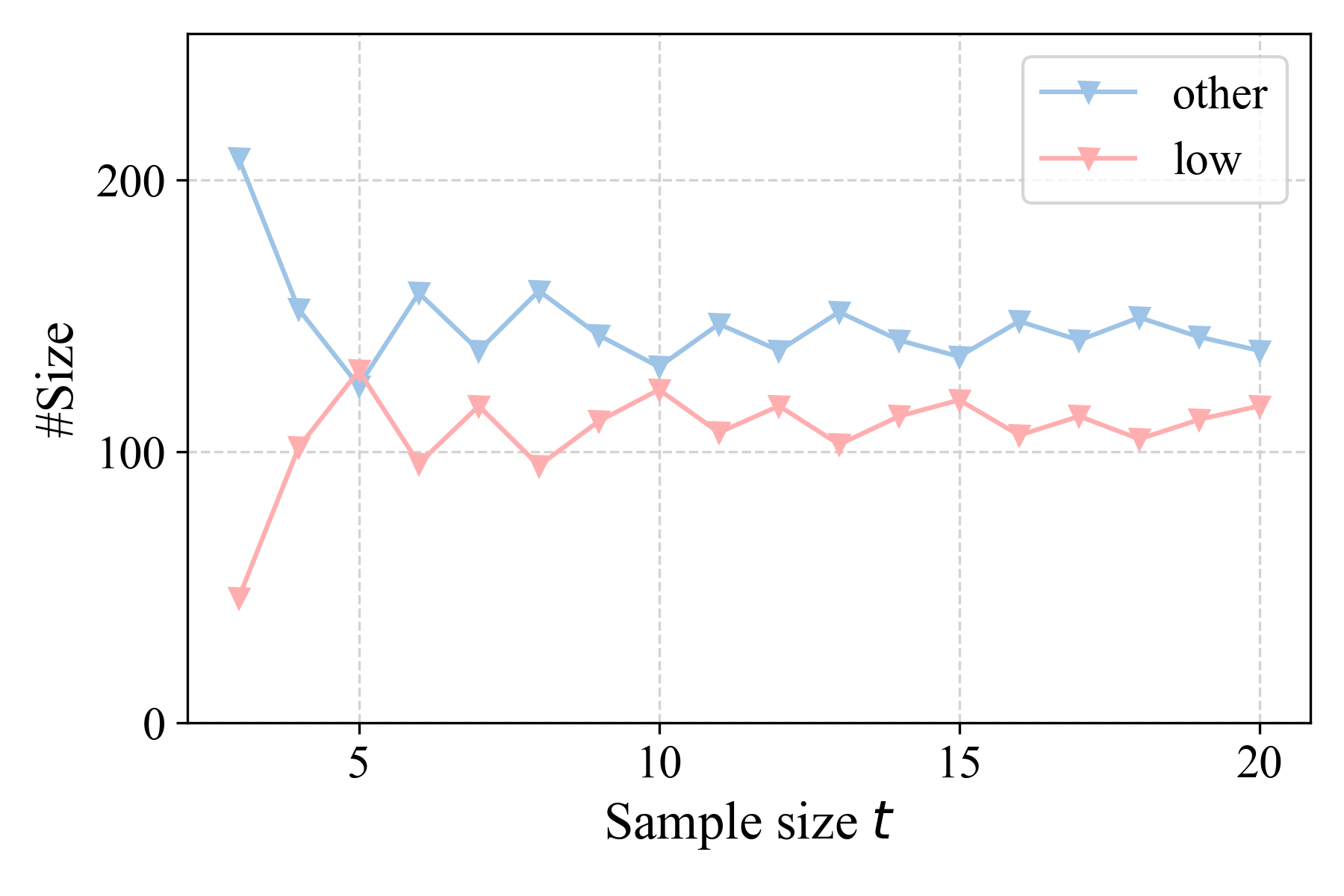}}
    \subfloat[Alg.]{\includegraphics[width=.33\textwidth]{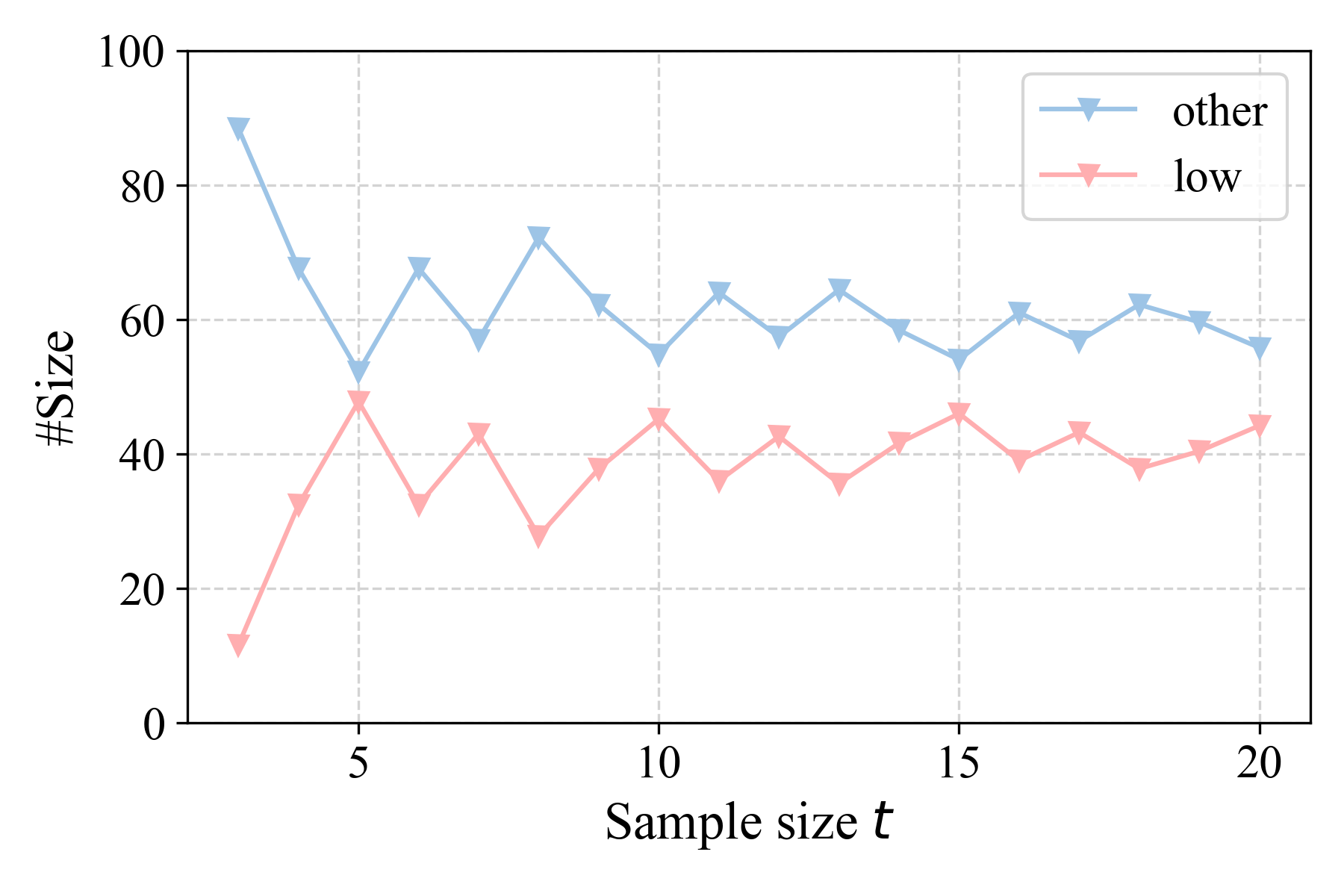}}
    \subfloat[Math.]{\includegraphics[width=.33\textwidth]{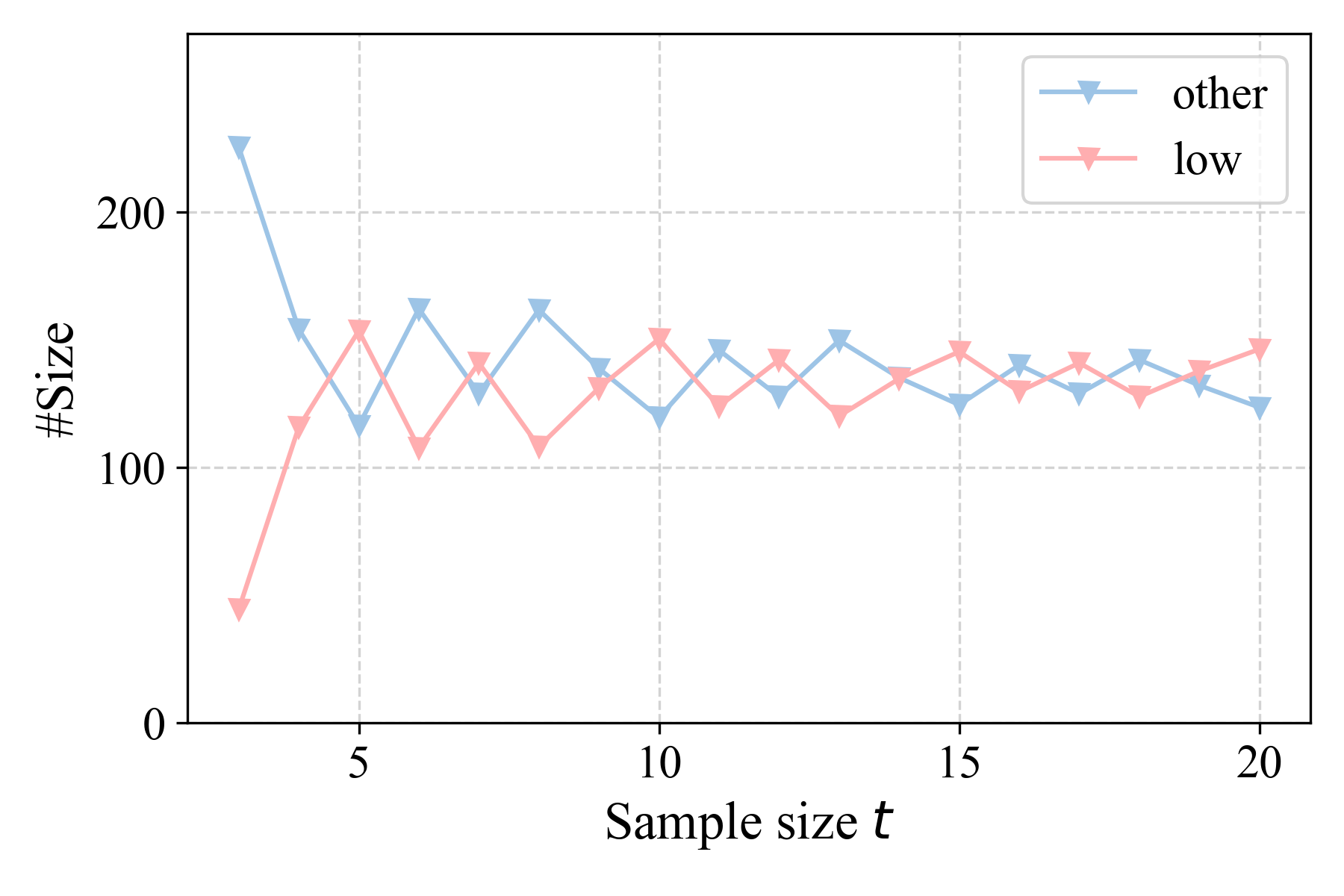}}\\
    \subfloat[CMS.]{\includegraphics[width=.33\textwidth]{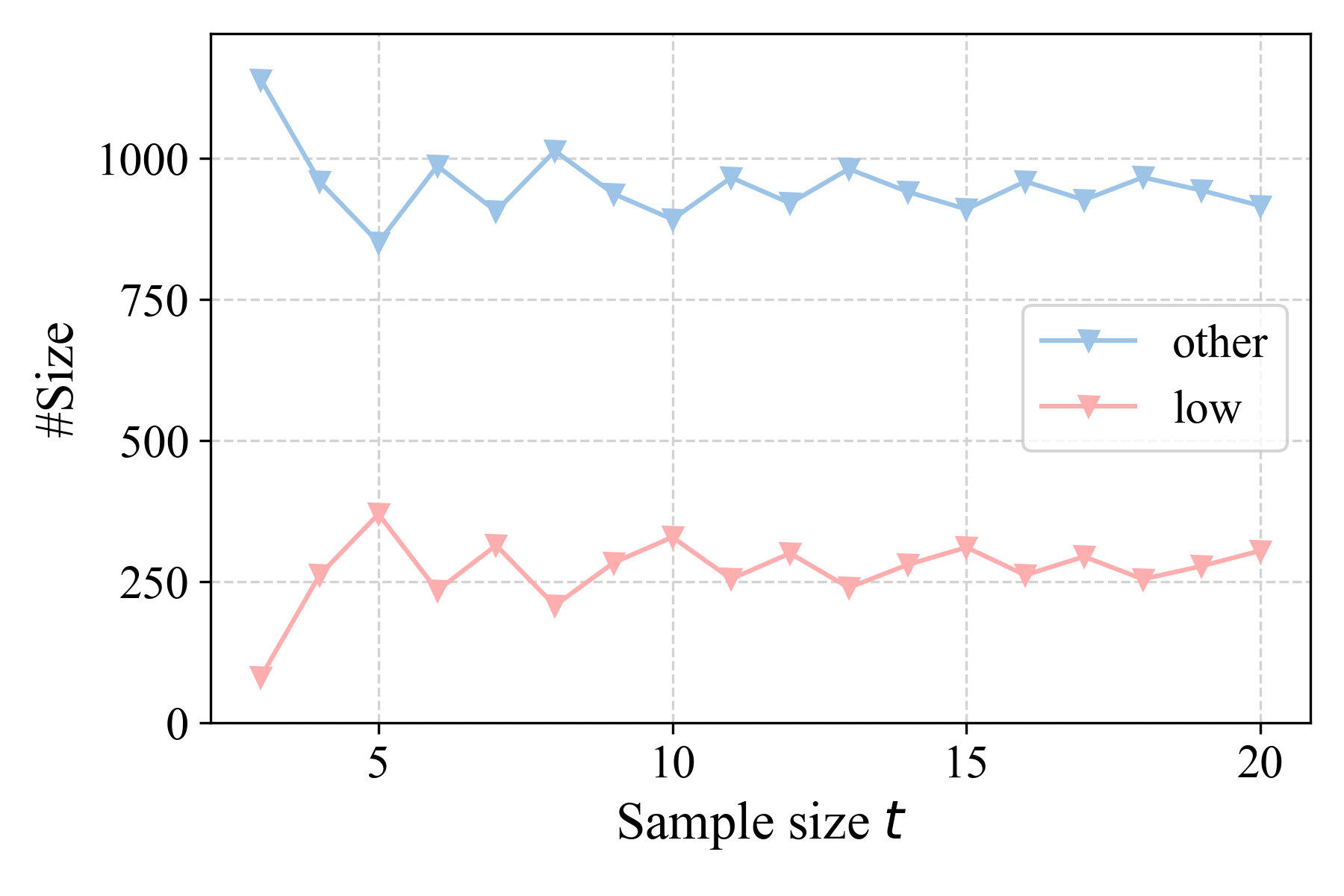}}
    \subfloat[OB.]{\includegraphics[width=.33\textwidth]{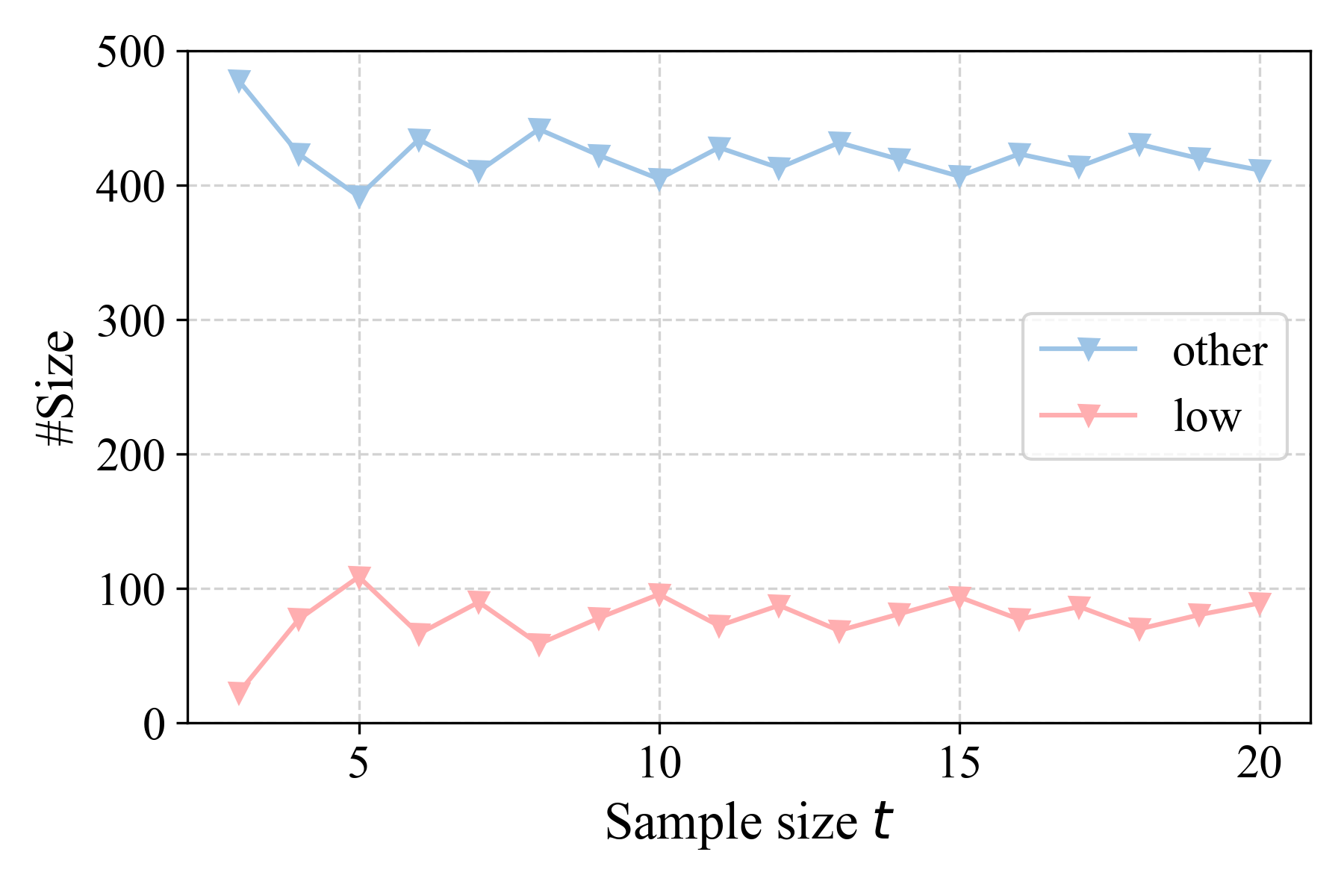}}
    \subfloat[ARC.]{\includegraphics[width=.33\textwidth]{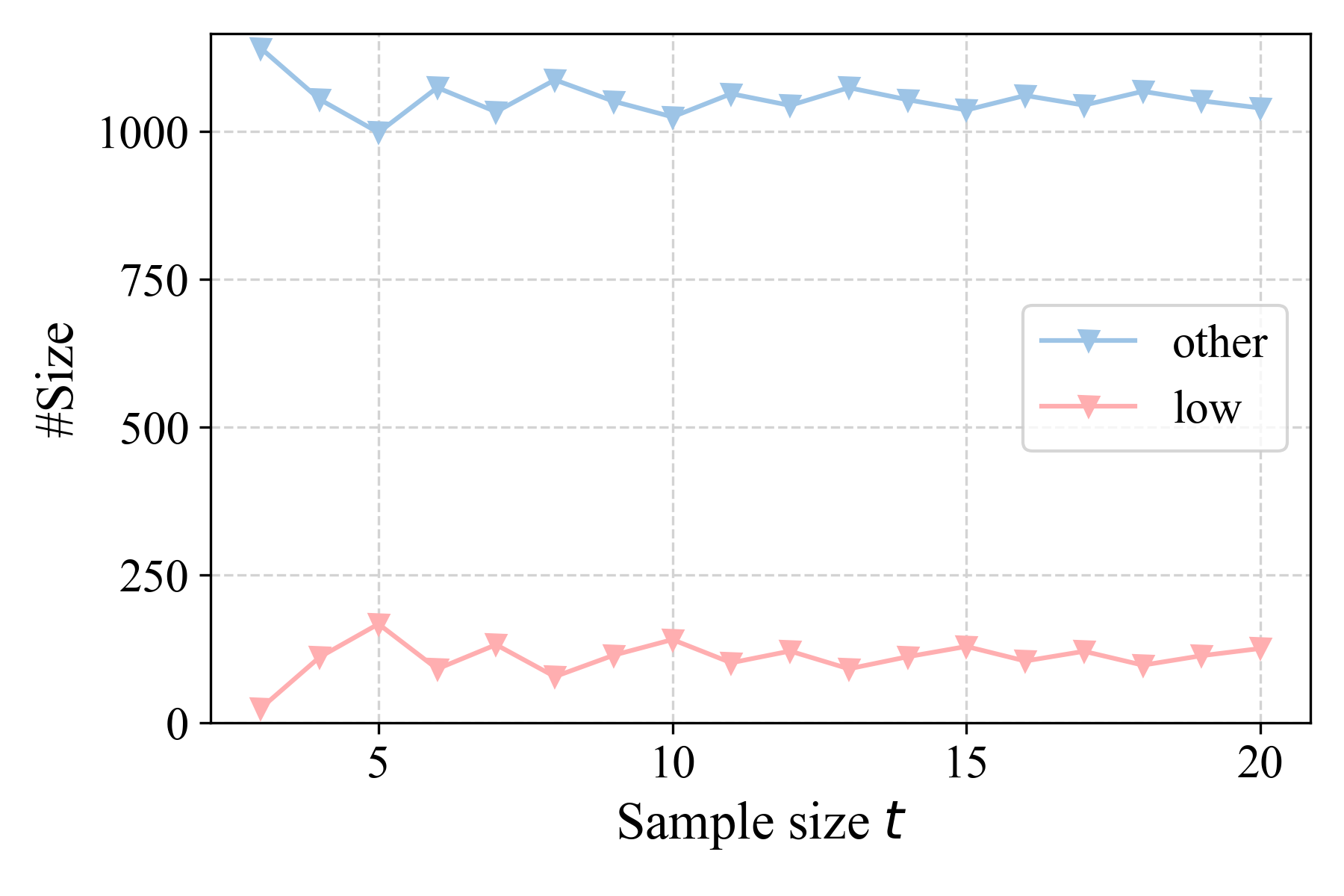}}\\
    \subfloat[Rid.]{\includegraphics[width=.33\textwidth]{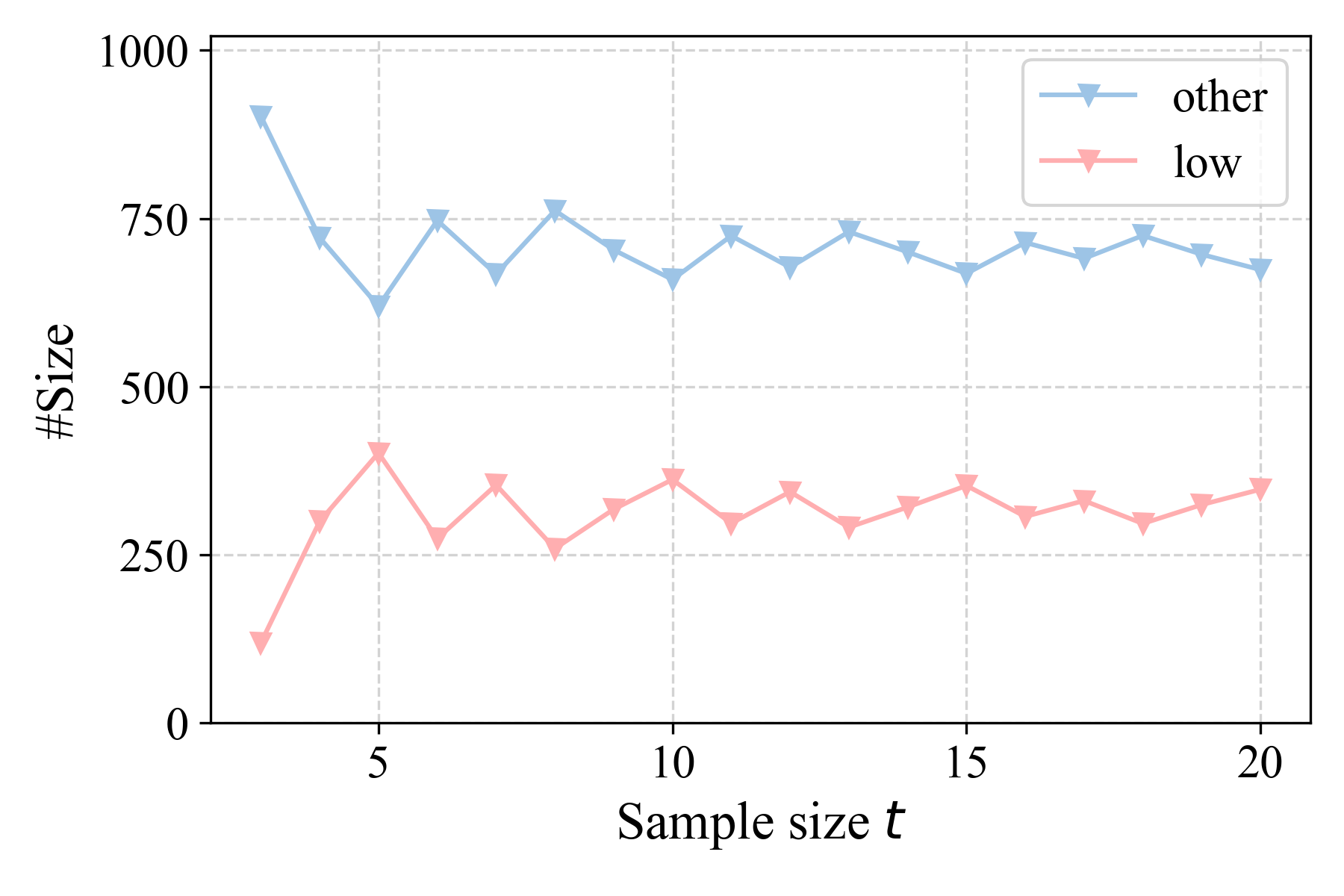}}
    \subfloat[Logi.]{\includegraphics[width=.33\textwidth]{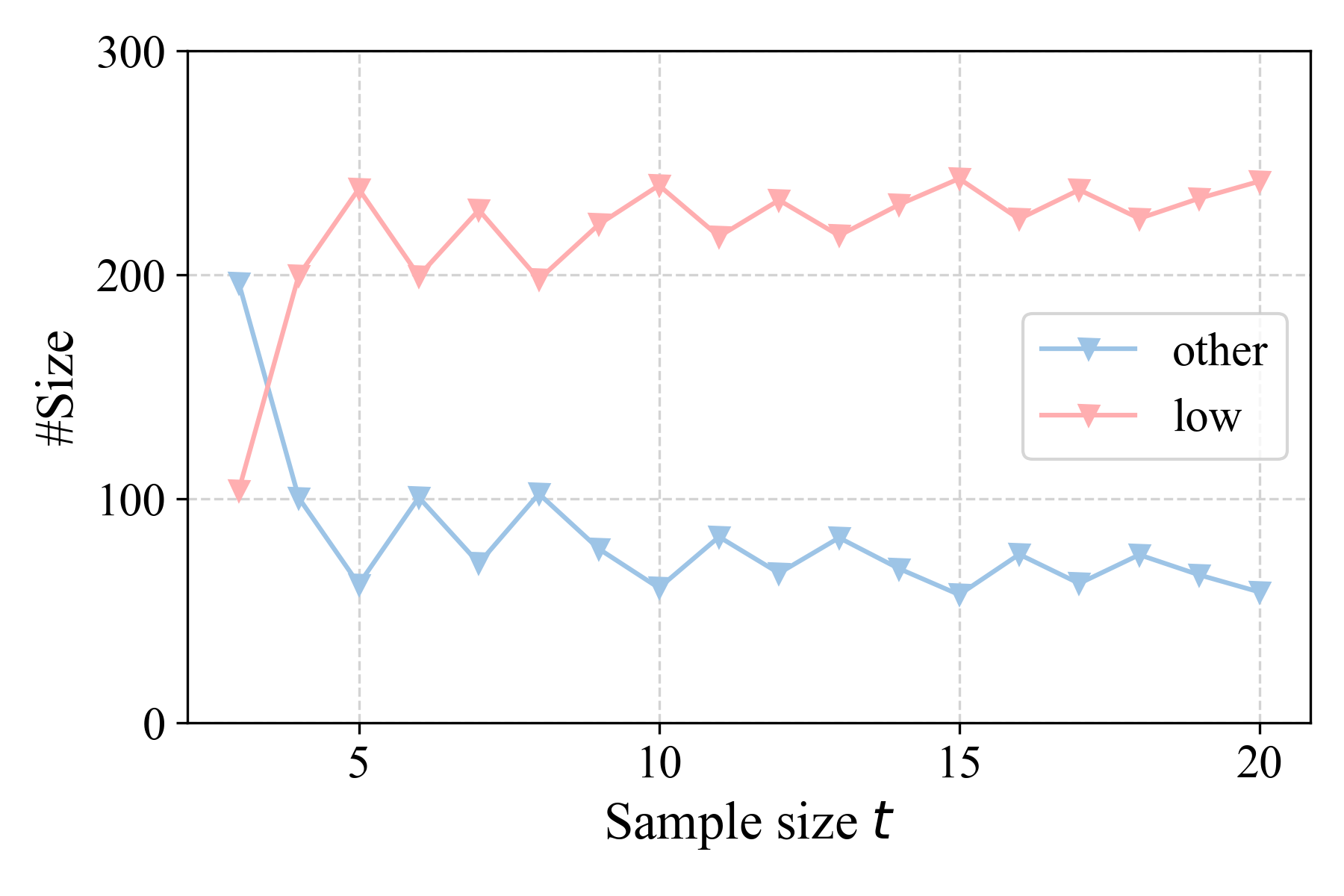}}
    \subfloat[Rec.]{\includegraphics[width=.33\textwidth]{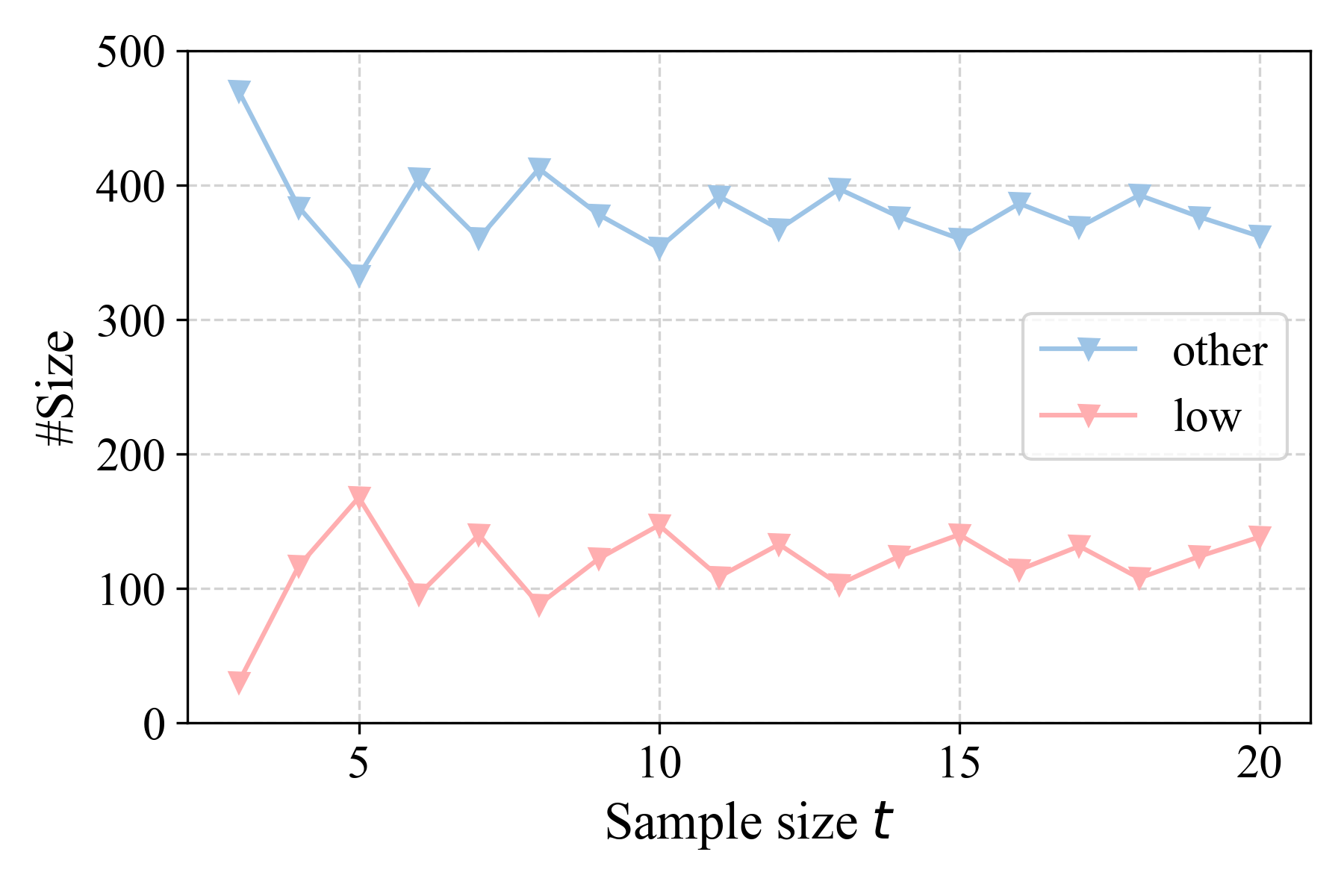}}
    \caption{The number of different subsets size for various sample size $t$ among each dataset.}
    \label{fig:scNumDivide_detail_size}
\end{figure*}

\end{document}